\pdfoutput=1
\documentclass[10pt,twocolumn,letterpaper]{article}

\usepackage[pagenumbers]{cvpr} % To force page numbers, e.g. for an arXiv version

\makeatletter
\@namedef{ver@everyshi.sty}{}
\makeatother

\usepackage{subfiles}
\usepackage{csquotes}
\usepackage{gensymb}
\usepackage{pifont}
\usepackage{textcomp}
\usepackage[table]{xcolor}
\usepackage{todonotes}
\usepackage{blindtext}
\usepackage{soul}
\usepackage{graphicx}
\usepackage{mathtools}
\usepackage{multirow}
\usepackage{flushend}
\usepackage{booktabs}
\usepackage{amsmath} 
\usepackage{amssymb} 
\usepackage{amsfonts}
\usepackage{bm} 
\usepackage{siunitx}
\usepackage{cancel}
\usepackage[inline]{enumitem}
\usepackage{microtype}
\usepackage{xcolor}
\usepackage{caption}
\usepackage[export]{adjustbox}
\usepackage{cuted}

\usepackage[pagebackref,breaklinks,colorlinks]{hyperref}

\usepackage[capitalize]{cleveref}
\crefname{section}{Sec.}{Secs.}
\Crefname{section}{Section}{Sections}
\Crefname{table}{Table}{Tables}
\crefname{table}{Tab.}{Tabs.}

\definecolor{mygreen}{HTML}{00A64F}
\definecolor{myred}{HTML}{ED1B23}

\newcommand{\net}{SkyEye}
\newcommand{\cmark}{\text{\ding{51}}}
\newcommand{\xmark}{\text{\ding{55}}}

\def\cvprPaperID{0000} % *** Enter the CVPR Paper ID here
\def\confName{CVPR}
\def\confYear{2023}

\begin{document}

%%%%%%%%% TITLE - PLEASE UPDATE
\title{SkyEye: Self-Supervised Bird's-Eye-View Semantic Mapping\\Using Monocular Frontal View Images}

\author{Nikhil Gosala$^{1}$\thanks{Equal contribution}
\hspace{1em} K\"ursat Petek$^{1*}$
\hspace{1em} Paulo L. J. Drews-Jr$^{1, 2}$
\hspace{1em} Wolfram Burgard$^{3}$
\hspace{1em} Abhinav Valada$^{1}$ \vspace{0.1cm} \\
$^1$University of Freiburg 
\hspace{0.65em} $^2$Federal University of Rio Grande
\hspace{0.65em} $^3$University of Technology Nuremberg \vspace{0.1cm}\\
\tt\small{\url{http://skyeye.cs.uni-freiburg.de}}}

\maketitle

%%%%%%%%% ABSTRACT
\begin{abstract}
Bird's-Eye-View~(BEV) semantic maps have become an essential component of automated driving pipelines due to the rich representation they provide for decision-making tasks.
However, existing approaches for generating these maps still follow a fully supervised training paradigm and hence rely on large amounts of annotated BEV data.
In this work, we address this limitation by proposing the first self-supervised approach for generating a BEV semantic map using a single monocular image from the frontal view~(FV). During training, we overcome the need for BEV ground truth annotations by leveraging the more easily available FV semantic annotations of video sequences. Thus, we propose the \net~architecture that learns based on two modes of self-supervision, namely, \textit{implicit supervision} and \textit{explicit supervision}. Implicit supervision trains the model by enforcing spatial consistency of the scene over time based on FV semantic sequences, while explicit supervision exploits BEV pseudolabels generated from FV semantic annotations and self-supervised depth estimates. Extensive evaluations on the KITTI-360 dataset demonstrate that our self-supervised approach performs on par with the state-of-the-art fully supervised methods and achieves competitive results using only 1\% of direct supervision in the BEV compared to fully supervised approaches. Finally, we publicly release both our code and the BEV datasets generated from the KITTI-360 and Waymo datasets.
\end{abstract}

%%%%%%%%% BODY TEXT
\section{Introduction}
\label{sec:introduction}

Bird's-Eye-View (BEV) maps are an integral part of an autonomous driving pipeline as they allow the vehicle to perceive the environment using a feature-rich yet computationally-efficient representation. These maps capture both static and dynamic obstacles in the scene while encoding their absolute distances in the metric scale using a low-cost 2D representation. Such characteristics allow them to be used in many distance-based time-sensitive applications such as trajectory estimation and collision avoidance~\cite{hurtado2021learning, honerkamp2021learning}. Existing approaches that estimate BEV maps from frontal view (FV) images and/or LiDAR scans require large datasets annotated in the BEV as they are trained in a fully supervised manner~\cite{cit:bev-seg-panopticbev, cit:bev-seg-bevfusion, cit:bev-seg-hdmapnet, younes2023catch}. However, BEV ground truth generation relies on the presence of HD maps, annotated 3D point clouds, and/or 3D bounding boxes, which are extremely arduous to obtain~\cite{meng2020weakly}. Recent approaches~\cite{cit:bev-seg-bevseg, cit:bev-seg-l2lao} circumvent this problem of requiring BEV ground truths by leveraging data from simulation environments. However, these approaches suffer from the large domain gap between simulated and real-world images, which results in their reduced performance in the real world.

\begin{figure}
    \centering
    \includegraphics[width=0.77\linewidth]{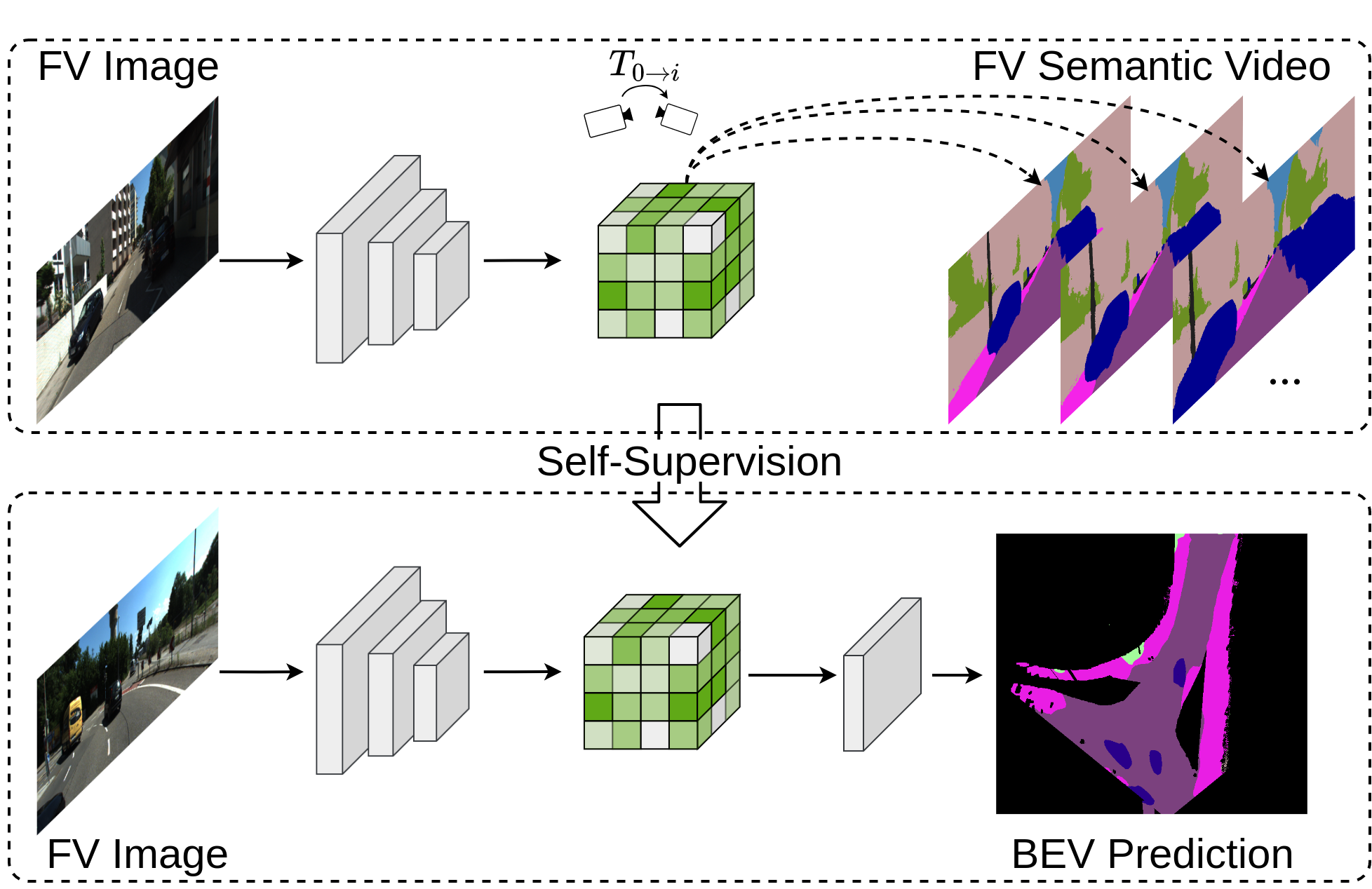}
    \vspace{-0.2cm}
    \caption{\net: The first self-supervised framework for semantic BEV mapping. We use sequences of FV semantic annotations to train the network to estimate a semantic map in BEV using a single RGB input.}
    \label{fig:teaser}
    \vspace{-0.5cm}
\end{figure}

% BEV map generation
In this work, we address the aforementioned limitations by proposing \textit{SkyEye}, the first self-supervised learning framework for generating an instantaneous semantic map in BEV, given a single monocular FV image. During training, our approach, depicted in \cref{fig:teaser}, overcomes the need for BEV ground truths by leveraging FV semantic ground truth labels along with the spatial and temporal consistency offered by video sequences. FV semantic ground truth labels can easily be obtained with reduced human annotation effort due to the relatively small domain gap between FV images of different datasets which allows for efficient label transfer~\cite{khoreva2017simple, shin2021labor, hurtado2022semantic}. Additionally, no range sensor is required for data recording. During inference, our model only uses a single monocular FV image to generate the semantic map in BEV.

Our proposed self-supervised learning framework leverages two supervision signals, namely, \textit{implicit} and \textit{explicit supervision}. Implicit supervision generates the training signal by enforcing spatial and temporal consistency of the scene. To this end, our model generates the FV semantic predictions for the current and future time steps using the FV image of only the current time step. These predictions are supervised using the corresponding ground truth labels in FV. Explicit supervision, in contrast, supervises the network using BEV semantic pseudolabels generated from FV semantic ground truths using a self-supervised depth estimation network augmented with a dedicated post-processing procedure.
We perform extensive evaluations of \textit{SkyEye} on the KITTI-360 dataset and demonstrate its generalizability on the Waymo dataset. Results demonstrate that \textit{SkyEye} performs on par with the state-of-the-art fully-supervised approaches and achieves competitive performance with only 1\% of pseudolabels in BEV. Further, we outperform all baseline methods w.r.t. generalization capabilities.

Our main contributions can thus be stated as follows:
\begin{itemize}[topsep=0pt]
\itemsep -0.15cm
    \item The first self-supervised framework for generating semantic BEV maps from monocular FV images.
    \item An implicit supervision strategy that leverages semantic annotations in FV to encode semantic and spatial information into a latent voxel grid.
    \item A pseudolabel generation pipeline to create BEV pseudolabels from FV semantic ground truth labels.
    \item A novel semantic BEV dataset derived from Waymo.
    \item Extensive evaluations as well as ablation studies to show the impact of our contributions.
    \item Publicly available code for our \textit{SkyEye} framework at \url{http://skyeye.cs.uni-freiburg.de}. 
\end{itemize}
\section{Related Work}
\label{sec:related-work}

In this section, we review the existing work related to BEV semantic mapping and self-supervised 3D representation learning based on monocular images.

{\parskip=5pt
\noindent\textbf{BEV Mapping}:
BEV map generation typically involves three stages: (i) FV feature extraction using an image encoder, (ii) feature transformation from FV to BEV, and (iii) BEV map generation using the transformed features - with most approaches focusing on FV-BEV transformation. The earliest approaches, VED~\cite{cit:bev-seg-lu2019ved} and VPN~\cite{cit:bev-seg-pan2020vpn} learn the FV-BEV mapping using a variational encoder-decoder architecture and a two-layer multi-layer perceptron respectively. However, they do not account for the geometry of the scene which results in their poor performance in the real world. Later approaches address this limitation by integrating scene geometry into the network design. PON~\cite{cit:bev-seg-pon} proposes an end-to-end network wherein a dense transformer module learns the mapping between a column in the FV image and a ray in the BEV prediction. LSS~\cite{cit:bev-seg-lss} uses a learnable categorical depth distribution to ``lift'' the FV features into the 3D space. Both these approaches, however, do not generalize across different semantic classes as the former employs a single mapping function throughout the image while the latter lacks the required network capacity.
PanopticBEV~\cite{cit:bev-seg-panopticbev} addresses these limitations by employing a dual-transformer approach to independently map the \textit{vertical} and \textit{flat} regions in the scene from FV to BEV. Recently, multiple approaches~\cite{cit:bev-seg-tiim, cit:bev-seg-crossviewtrans, cit:bev-seg-vitbevseg} have proposed using vision transformer-based architectures to learn the FV-BEV mapping, while others have explored incorporating range sensors such as LiDARs and Radars into the BEV map generation pipeline~\cite{cit:bev-seg-bevfusion, cit:bev-seg-hdmapnet}. It is important to note that all the aforementioned approaches follow a fully-supervised training strategy and hence rely on BEV ground truth labels for training. Although such approaches result in state-of-the-art performance, their reliance on BEV ground truth labels severely impacts their scalability.
In this paper, we propose the first self-supervised approach using only FV image sequences, their corresponding FV semantic annotations, and the ego-poses for training.} 

{\parskip=5pt
\noindent\textbf{Self-Supervised Monocular 3D Representation Learning}:
This task forms one of the fundamental challenges of computer vision and has found use in applications such as novel view synthesis and 3D scene reconstruction~\cite{bevsic2022dynamic}. Early works use geometry-based approaches such as structure-from-motion~\cite{schonberger2016structure}, multi-view stereo~\cite{furukawa2009accurate}, and multi-hypothesis labeling~\cite{hoiem2005geometric}, while recent approaches typically employ deep learning-based solutions~\cite{katircioglu2018learning, vodisch2022continual} to address this challenge. More recently, self-supervised approaches have been proposed to alleviate the amount of annotated data required to learn the 3D structure. Video Autoencoder~\cite{lai2021video} uses an autoencoder to learn the 3D structure of a static scene for the task of novel view synthesis. In the context of robotics, self-supervised representation learning has been used for tasks such as depth estimation~\cite{cit:monodepth2-godard, guizilini20203d}, surface normal estimation~\cite{goyal2019scaling}, optical flow~\cite{liu2019selflow, liu2020flow2stereo}, visual-inertial odometry~\cite{Han19}, keypoints estimation~\cite{von2022self}, stereo matching~\cite{yang2021self}, image enhancement ~\cite{mello2022underwater}, and scene flow~\cite{hur2020self} among many others. These approaches have shown tremendous potential in the real world due to their ability to efficiently scale across multiple locations without needing expensive human intervention. We extend this set of self-supervised approaches by proposing the first self-supervised learning framework to predict BEV semantic maps without the need for any BEV ground truth data.}
\section{Technical Approach}
\label{sec:technical-approach}

\begin{figure*}
    \centering
    \includegraphics[width=0.87\textwidth]{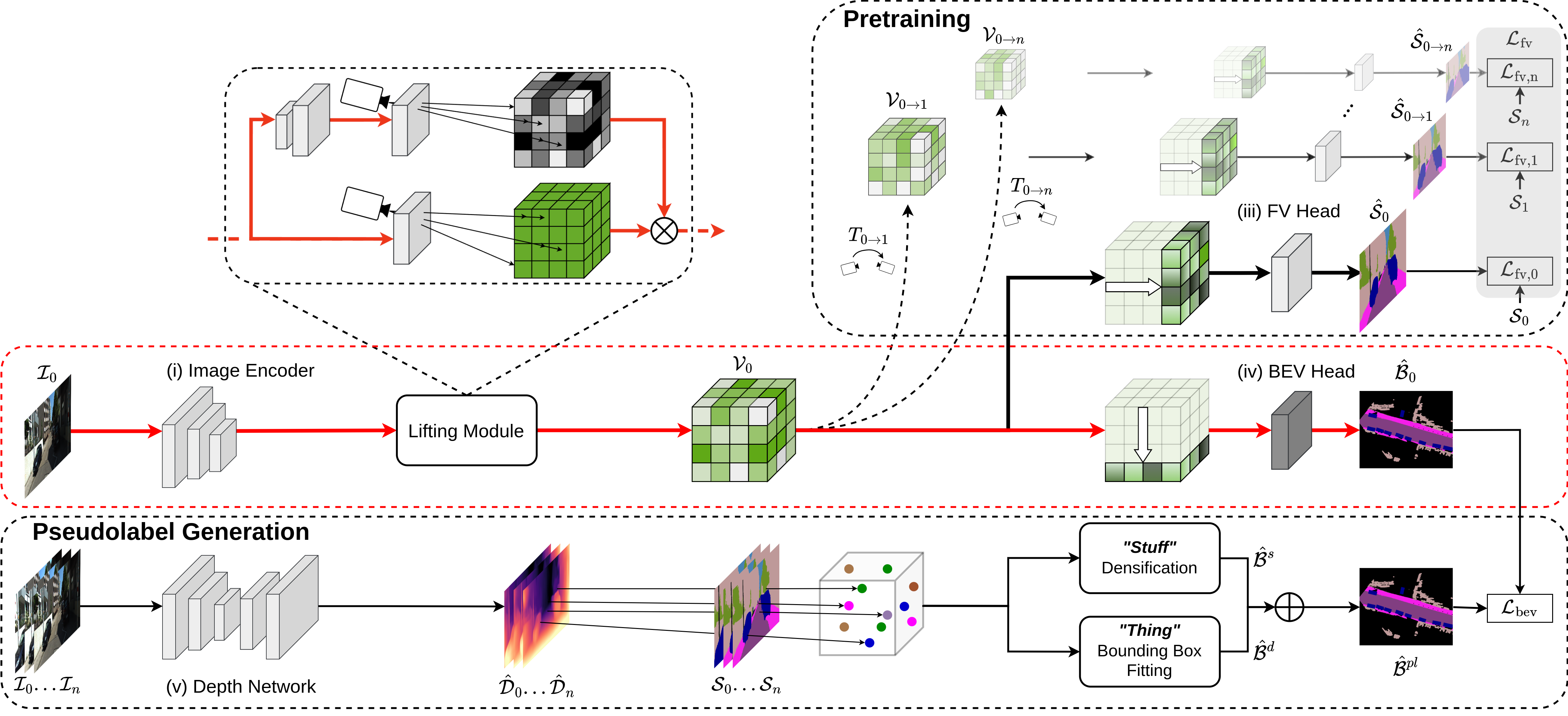}
    \vspace{-0.2cm}
    \caption{Overview of our proposed self-supervised BEV semantic mapping framework, \net. The core component of our approach is the latent voxel grid $\mathcal{V}_0$ that serves as a joint feature representation for segmentation tasks in FV and BEV. We encode spatial and semantic information into the voxel grid using \textit{implicit supervision} during a pretraining step and \textit{explicit supervision} in a subsequent refinement step using pseudolabels that are generated with a self-supervised depth prediction pipeline. The path in red denotes the processing steps during inference time.}
    \label{fig:arch-network}
    \vspace{-0.2cm}
\end{figure*}

In this section, we present our novel self-supervised learning framework, \net, for generating BEV semantic maps from a single monocular FV image without any ground truth supervision in BEV. The core idea of our approach is to generate an intermediate 3D voxel grid that serves as a joint feature representation for both FV and BEV segmentation tasks, thus allowing us to leverage FV supervision to augment the BEV semantic learning procedure. An overview of our proposed self-supervised pipeline is depicted in \cref{fig:arch-network}.

Our framework generates the supervision signal using two strategies, namely, \textit{implicit supervision} and \textit{explicit supervision}. Implicit supervision generates the training signal by exploiting the spatial and temporal consistency of the scene via FV Semantic Scene Consistency ($\mathcal{L}_{fv}$, \cref{subsec:implicit-supervision}) which operates on the depth-wise projection of the voxel grid.
The explicit supervision, in turn, operates on the orthographic height-wise projection of the voxel grid and provides supervision in BEV via pseudolabels generated in a self-supervised manner ($\mathcal{L}_{bev}$, \cref{subsec:explicit-supervision}).
The final loss is thus computed as:
\begin{equation}
    \mathcal{L} = \mathcal{L}_{fv} + \mathcal{L}_{bev}
\end{equation}

In the following sections, we present an overview of our network architecture and provide further insight into the computation of the aforementioned losses. Further, in all the upcoming notations, the subscript $i$ refers to an instance of an element at time step $t_i$.

\subsection{Network Architecture}
\label{subsec:network-architecture}

Our model comprises five major components: (i)~an image encoder to generate 2D image features, (ii) a lifting module to generate the 3D voxel grid using a learned depth distribution, (iii) an FV semantic head to generate the FV semantic predictions for implicit supervision, (iv)~a BEV semantic head to generate the BEV semantic map, and (v)~an independent self-supervised depth network to generate the BEV pseudolabels. 
\cref{fig:arch-network} presents an overview of our proposed framework. \\
The encoder follows the EfficientDet-D3 backbone~\cite{tan2020efficientdet} which takes an FV image as input and outputs 2D features at four different scales which we subsequently merge using the multi-scale fusion strategy outlined in EfficientPS~\cite{cit:fv-efficientps}.
The lifting module projects the 2D features to a 3D voxel grid representation using the camera projection equation coupled with a learned depth distribution that provides the likelihood of features in a given voxel.
We then process the voxel grid depth-wise or height-wise based on whether the output is in the FV or BEV respectively. We generate the FV semantic logits by applying perspective distortion to the 3D voxel grid, flattening it along the depth dimension, and mapping it to the output channels using a $1\times1$ convolution.
Similarly, we generate the BEV logits by flattening the 3D voxel grid orthographically along the height dimension and passing it through a $1\times1$ convolution to generate the required output channels in the BEV.
Our self-supervised depth network is independent from the aforementioned model and is only used to generate the BEV semantic pseudolabels. It uses a separate instance of EfficientDet-D3 backbone and feature merging module outlined above. The depth decoder consists of three upsampling layers, each of which follow the upsampling strategy defined in~\cite{cit:monodepth2-godard}. The final depth is then computed by applying a $3\times3$ convolution, normalizing it using a sigmoid function and scaling it to the required range.
\subsection{Implicit Supervision}
\label{subsec:implicit-supervision}
\begin{figure*}
    \centering
    \includegraphics[width=0.85\linewidth]{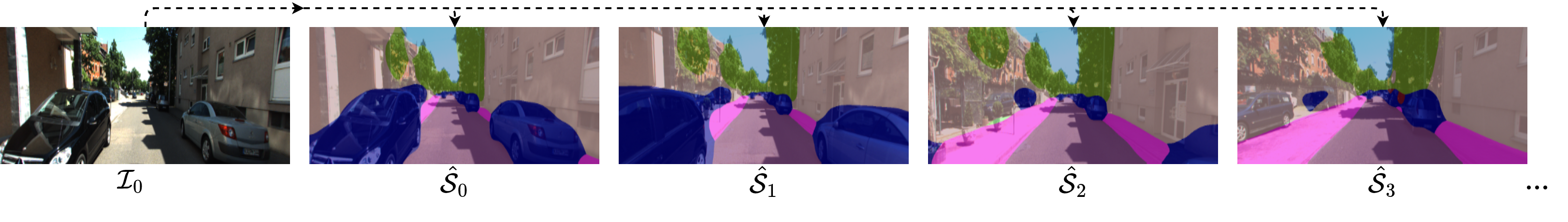}
    \vspace{-0.3cm}
    \caption{Semantic predictions of SkyEye for future time steps using the FV image of only the initial time step. The disocclusion of sidewalks in the semantic predictions indicates that SkyEye can reason about both occluded regions and spatial extents of objects in the scene with the encoded semantic information.}
    \label{fig:fvsempred-warping}
\end{figure*}

Autonomous driving scenes comprise many static elements such as parked cars and buildings which establish a strong framework for generating a supervision signal by exploiting their consistency over multiple time steps. We exploit this characteristic of the real world and generate the implicit supervision signal by enforcing consistency between FV semantic predictions at multiple time steps. To this end, we predict the FV semantic maps for the initial ($t_0$) as well as future time steps ($t_1, ..., t_n$) using only the intermediate voxel grid representation at the initial time step as depicted in \cref{fig:fvsempred-warping}. We hypothesize that this formulation would help the network generate a spatially consistent volumetric representation of the scene from a single FV image. Further, we also  hypothesize that this formulation would help the voxel grid encode complementary information from multiple images to resolve occlusions and hence play a pivotal role in generating accurate BEV semantic maps from the limited view of only a single time step.

We first use the provided FV monocular image $\mathcal{I}_0$ to generate the intermediate 3D voxel grid representation $\mathcal{V}_0$. 
This voxel grid is perspectively distorted using the camera intrinsics and processed along the depth dimension to generate the FV semantic prediction $\mathcal{\hat{S}}_0$.
Perspective distortion of the voxel grid prior to processing it in FV is crucial to prevent implicit supervision from distorting the voxel grid.
Parallelly, we transform $\mathcal{V}_0$ to generate the voxel grids $\mathcal{V}_{0\rightarrow i} = T_{0\rightarrow i} \mathcal{V}_{0}$ for future time steps $(t_1,...t_n)$ using the relative ego poses $T_{0\rightarrow i}$ between the initial and future time steps. We then use the generated pseudo voxel grids to infer the future FV semantic predictions $\mathcal{\hat{S}}_{0\rightarrow1}, \mathcal{\hat{S}}_{0\rightarrow2}, ..., \mathcal{\hat{S}}_{0\rightarrow n}$.
Subsequently, we compute the cross entropy loss between the FV semantic predictions and their corresponding FV semantic ground truths to generate the implicit supervision signal for training the model. We linearly down-weight the loss for future time steps to negate the ill effects of dynamic objects and error propagation during model training. Thus, we compute the FV semantic scene consistency loss $\mathcal{L}_{fv}$ by accumulating the losses for each time step $\mathcal{L}_{fv,i}$ as:
\begin{equation}
    \mathcal{L}_{fv} = \sum_{i=0}^{n} L_{fv,i} = \sum_{i=0}^{n} w_i CE(\mathcal{\hat{S}}_{0\rightarrow i}, \mathcal{S}_i),
\end{equation}
where $w_i$ refers to the time step-based weight which linearly decays from $1$ to $0.2$, and $CE(a, b)$ refers to the cross entropy loss between tensors $a$ and $b$.

\subsection{Explicit Supervision}
\label{subsec:explicit-supervision}
Our model comprises a BEV segmentation head with learnable parameters that is designed to generate the desired BEV semantic map.
However, implicit supervision does not generate a gradient flow through the BEV head, which underlines the need for \textit{explicit supervision} in BEV. To this end, we propose a pseudolabel generation procedure that consists of three steps as depicted in \cref{fig:arch-network}, namely (i) a depth prediction pipeline to lift FV semantic annotations into BEV yielding a semantic point cloud, (ii) an instance generation module based on DBSCAN~\cite{schubert2017dbscan} to incorporate prior knowledge on dynamic objects, and (iii) a densification module to generate dense segmentation masks from sparse depth predictions for static classes.
\begin{figure*}
    \centering
    \includegraphics[width=0.85\linewidth]{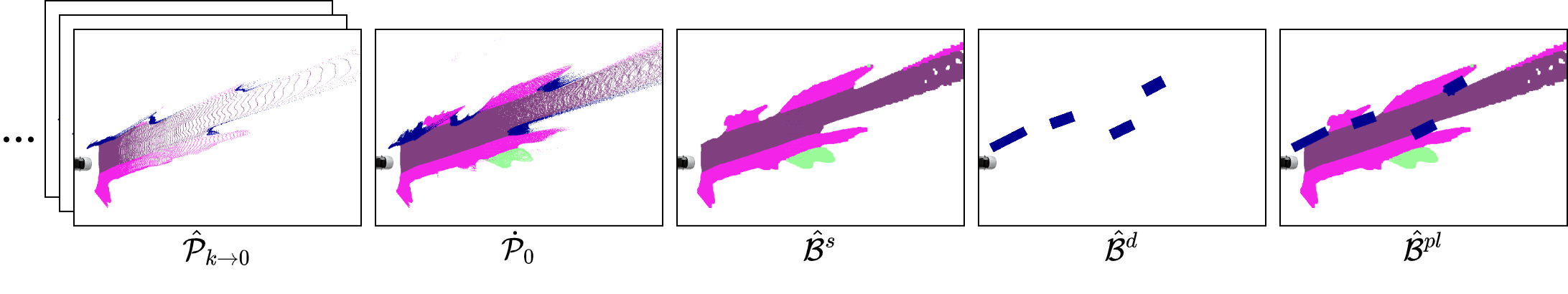}
    \vspace{-0.5cm}
    \caption{Overview of our pseudolabel generation pipeline. We lift semantic annotations in FV into the 3D world ($\mathcal{\hat{P}}_{k\rightarrow 0}$) and accumulate them ($\dot{\mathcal{P}}_0$). We then densify the static classes in BEV ($\hat{\mathcal{B}}^s$) and fit bounding boxes around clustered dynamic objects ($\hat{\mathcal{B}}^d$). $\hat{\mathcal{B}}^s$ and $\hat{\mathcal{B}}^d$ are then merged to generate the final semantic pseudolabel map $\hat{\mathcal{B}}^{pl}$.}
    \label{fig:explicit-supervision-overview}
    \vspace{-0.3cm}
\end{figure*}
Prior to pseudolabel generation, we train our depth network on the corresponding dataset in a self-supervised manner as proposed in~\cite{cit:monodepth2-godard} using the ego poses to ensure metric scale of the depth estimates. 

{\parskip=5pt
\noindent\textbf{Pseudolabel Generation}:
We generate pseudolabels for time step $t_0$ by employing a sequence $\mathcal{W}$ of FV images and their corresponding semantic ground truths. \cref{fig:explicit-supervision-overview} shows an example of each proposed step.
We first predict the depth map $\hat{\mathcal{D}_{i}}$ for each FV image $\mathcal{I}_{i}\in \mathcal{W}$ to lift FV semantic ground truths into BEV using the known camera intrinsics and poses. We then accumulate the semantic point clouds and transform them into the perspective of $t_0$, to obtain a single accumulated semantic point cloud $\dot{\mathcal{P}_0} = \bigcup_{k \in \mathcal{W}} \hat{\mathcal{P}}_{k \rightarrow 0}$.}
For dynamic objects, we retain only those points in the point cloud that are consistent with the FV semantic ground truth at $t_0$ to prevent object motion from corrupting the supervision signal.
We then use the accumulated point cloud to both create dense semantic labels for static classes and fit bounding boxes around dynamic objects (\cref{fig:arch-network}).\\
Firstly, we orthographically project points belonging to static classes to generate a sparse BEV map. 
We then densify the sparse BEV map by applying a series of morphological dilate and erode operations to generate the first set of labels $\hat{\mathcal{B}^s}$ for static classes.
Secondly, we try to mitigate the lack of observability of the shape of dynamic objects by incorporating prior knowledge and fitting bounding boxes around each object instance. However, we are faced with two challenges: (i) the FV data has no notion of object instances, and (ii) the predicted depth maps are prone to outliers due to the presence of transparent and reflective surfaces. We address these challenges by introducing the notion of instances and rejecting outliers in depth maps. We do so by clustering the accumulated semantic point cloud using DBSCAN to yield a set $\mathcal{M}$ of $C$ clusters, where $\mathcal{M} = \{m_j, j = 1...C\}$. The points belonging to the $M$ clusters are then orthographically projected into the BEV space and an ellipse $\mathcal{E} = \{x_c, y_c, a, b, \theta\}$, which serves as an analytical and differentiable replacement of a bounding box, is fit around each cluster using the RANSAC~\cite{cit:bev-seg-ransac} algorithm. The predicted ellipse parameters with its 2D center point ($x_c$, $y_c$), semi-minor axes ($a$, $b$) and orientation $\theta$ define the position, extents, and orientation of the bounding box in BEV, respectively. 
This procedure generates a second BEV map $\mathcal{\hat{B}}^d$ which is then overlaid on $\mathcal{\hat{B}}^s$ to generate the final pseudolabel map $\mathcal{\hat{B}}^{pl}$ containing both static and dynamic classes.
Finally, this pseudolabel map is used to further supervise the semantic BEV map at the network output using the cross-entropy loss as:
\begin{equation}
    \mathcal{L}_{bev} = CE(\mathcal{\hat{B}}^{pl}, \hat{\mathcal{B}}).
\end{equation}
\section{Experimental Results}
\label{sec:experiments}

In this section, we present the quantitative and qualitative results of our proposed self-supervised BEV semantic map generation pipeline, \net, along with comprehensive ablation studies to highlight the importance of our contributions. We also present the datasets used for experimental evaluation and provide a detailed description of the training protocol to ensure transparency and result reproduction.

\subsection{Datasets}
We evaluate \net~on the KITTI-360~\cite{cit:dataset-kitti360} dataset and study its generalization ability by pre-training it on KITTI-360 and evaluating it on Waymo Open Dataset~\cite{cit:dataset-waymo}.
We select these datasets to evaluate our approach on a wide variety of driving scenarios encountered in different regions of the world. Since neither KITTI-360 nor Waymo provide BEV semantic labels, we follow the data generation process outlined in PanopticBEV~\cite{cit:bev-seg-panopticbev} to generate the BEV semantic ground truth labels. We slightly modify this process and remove the occlusion masking step to make BEV labels occlusion-agnostic. It is important to note that the generated BEV ground truths are only used to train the fully-supervised baselines and perform the quantitative evaluation, and are \textit{not} used in our self-supervised learning framework. 
Of the $10$ sequences in the KITTI-360 dataset, we use sequence $10$ for validation and use the remaining sequences for training. For the generalization experiment, we evaluate our pre-trained model on all samples in the Waymo validation split. 

% Semantic Evaluation
\begin{table*}
\scriptsize
\centering
\caption{Evaluation of BEV semantic mapping on the KITTI-360 dataset. All metrics are reported in $[\%]$.}
\vspace{-0.2cm}
\label{tab:quant-eval-kitti}
\setlength\tabcolsep{3.7pt}
 \begin{tabular}{l|c|cccccccc|c}
 \toprule
\textbf{Method} & \textbf{BEV GT} & \textbf{Road} & \textbf{Sidewalk} & \textbf{Building} & \textbf{Terrain} & \textbf{Person} & \textbf{2-Wheeler} & \textbf{Car} & \textbf{Truck} & \textbf{mIoU}
 \\
 \midrule
IPM~\cite{cit:bev-seg-ipm-original} & \xmark &53.03 & 24.90 & 15.19 & 32.31 & 0.20 & 0.36 & 11.59 & 1.90 & 17.44 \\
TIIM~\cite{cit:bev-seg-tiim} & \cmark &63.08 & 28.66 & 13.70 & 25.94 & 0.56 & \textbf{6.45} & 33.31 & 8.52 & 22.53 \\
VED~\cite{cit:bev-seg-lu2019ved} & \cmark & 65.97 & 35.41 & 37.28 & 34.34 & 0.13 & 0.07 & 23.83 & 8.89 & 25.74 \\
VPN~\cite{cit:bev-seg-pan2020vpn} & \cmark & 69.90 & 34.31 & 33.65 & 40.17 & 0.56 & 2.26 & 27.76 & 6.10 & 26.84 \\
PON~\cite{cit:bev-seg-pon} & \cmark & 67.98 & 31.13 & 29.81 & 34.28 & 2.28 & 2.16 & 37.99 & 8.10 & 26.72 \\
PoBEV~\cite{cit:bev-seg-panopticbev} & \cmark & 70.14 & 35.23 & 34.68 & 40.72 & 2.85 & 5.63 & \textbf{39.77} & \textbf{14.38} & 30.42 \\
\cmidrule{1-11}
\net~(Ours) & \xmark & \textbf{71.39} & \textbf{37.62} & \textbf{37.48} & \textbf{44.38} & \textbf{4.73} & 4.72 & 32.73 & 10.84 & \textbf{30.49} \\
\bottomrule
\end{tabular}
\vspace{-0.3cm}
\end{table*}

\subsection{Training Protocol}
We train \net~on images of size $1408\times384$ pixels by following a two-step training protocol. First, we learn to infer the 3D geometry of the scene from a single FV image by training the model using only implicit supervision on a window size of $10$ for $20$ epochs with a learning rate~(LR) of $0.005$. We sample every second image from the window to capture a long time horizon while reducing the training time of the model. We then specialize the model for BEV segmentation by explicitly supervising it using the generated BEV pseudolabels for $20$ epochs and LR of $0.001$. We augment the dataset during both stages using random combinations of horizontal image flipping, as well as color perturbations via changes to image brightness, contrast, and saturation. We optimize the network across both training steps using SGD with a batch size of $12$, momentum of $0.9$, and weight decay of $0.0001$. We follow a multi-step training schedule wherein we decay LR by a factor of $0.5$ at epoch $15$ and $0.2$ at epoch $18$. We initialize the EfficientDet backbone using weights from COCO pretraining while the other layers are initialized using the Kaiming-He~\cite{he2015delving} initialization strategy.

\subsection{Quantitative Results}

Since we are the first to propose a method for self-supervised BEV segmentation, we benchmark \net~against EfficientPS~\cite{cit:fv-efficientps} + IPM~\cite{cit:bev-seg-ipm-original} as well as $5$ fully-supervised approaches, namely Translating Images Into Maps (TIIM)~\cite{cit:bev-seg-tiim}, Variational Encoder Decoder (VED)~\cite{cit:bev-seg-lu2019ved}, View Parsing Network (VPN)~\cite{cit:bev-seg-pan2020vpn}, Pyramid Occupancy Network (PON)~\cite{cit:bev-seg-pon}, and PanopticBEV (PoBEV)~\cite{cit:bev-seg-panopticbev}. We train the baseline models on our dataset using the open source code provided by the authors after minimally adapting them to handle the different input size, output size, and number of output semantic classes. We ensure fair comparison by adhering as closely as possible to the training protocols outlined in their respective publications. \cref{tab:quant-eval-kitti} presents the results of this evaluation using the class-wise Intersection-over-Union (IoU) and overall mean IoU (mIoU) metrics for the KITTI-360 dataset. 

We observe that our proposed \net~model outperforms $5$ of $6$ baseline models by more than \SI{3.65}{pp} and performs on par with the state-of-the-art fully-supervised approach PoBEV while not using any form of BEV ground truth supervision. We further note that our approach exceeds the baselines by up to \SI{3.66}{pp} on all static classes such as \textit{road}, \textit{sidewalk}, \textit{building}, and \textit{terrain}. This superior performance for static classes can largely be attributed to our consistency-based implicit supervision which enables the network to infer spatially consistent features as well as reason about occluded and distant regions. At the same time, we observe that our model demonstrates inferior performance on dynamic classes such as \textit{car} and \textit{truck} as compared to PoBEV, reporting nearly \SI{7}{pp} and \SI{3.5}{pp} lower on \textit{car} and \textit{truck} respectively. This result is caused by the use of only a forward-facing FV image which impedes the network from reasoning about the shape and extent of various objects. Further, the sparsity of points on dynamic objects in distant regions hinders the generation of accurate pseudolabels and negatively impacts the performance of our pipeline. Nevertheless, our approach still extracts strong features for dynamic objects from both implicit and explicit supervision which allows it to outperform IPM, VED, and VPN, and be on par with TIIM.

%%%%%%%%%%%%%%%%%%%%%%%%%%%%%%%%%% ABLATIVE EXPERIMENTS %%%%%%%%%%%%%%%%%%%%%%%%%%%%%%%%%%
\subsection{Ablation Study}

\begin{table*}
\centering
\caption{Ablation study on the impact of Implicit Supervision on the overall network performance. All scores are reported on the KITTI-360 dataset.}
\vspace{-0.2cm}
\label{tab:ablation-implicit-percentages}
\scriptsize
\setlength\tabcolsep{3.7pt}
 \begin{tabular}{c|ccc|c|cccccccc|c}
 \toprule
 \textbf{BEV~(\%)} & \textbf{Method} & \textbf{BEV GT} & \textbf{Implicit} & \textbf{Epochs} & \textbf{Road} & \textbf{Sidewalk} & \textbf{Building} & \textbf{Terrain} & \textbf{Person} & \textbf{2-Wheeler} & \textbf{Car} & \textbf{Truck} & \textbf{mIoU} \\
 \midrule
 \multirow{3}{*}{0.1} & PoBEV & \cmark & - & \multirow{3}{*}{300} & 54.73 & 19.08 & 22.63 & 5.40 & 0.00 & 0.00 & 3.81 & 0.00 & 13.21 \\
 & \net & \xmark & \xmark & & 55.20 & 18.42 & 20.95 & 11.63 & 0.00 & 0.00 & 15.53 & 0.00 & 15.22 \\
 & \net & \xmark & \cmark & & \textbf{68.49} & \textbf{31.11} & \textbf{32.98} & \textbf{29.92} & 0.00 & 0.00 & \textbf{19.19} & 0.00 & \textbf{22.71} \\
 \midrule
 \multirow{3}{*}{1} & PoBEV & \cmark & - & \multirow{3}{*}{100} & 61.70 & 17.10 & 27.81 & 26.72 & 0.07 & 0.36 & 21.51 & 0.84 & 19.51 \\
 & \net & \xmark & \xmark & & 64.25 & 22.43 & 32.20 & 24.41 & 0.44 & 0.00 & 24.09 & 0.69 & 21.06 \\
 & \net & \xmark & \cmark & & \textbf{72.00} & \textbf{33.76} & \textbf{37.59} & \textbf{38.75} & \textbf{3.77} & \textbf{1.81} & \textbf{28.04} & \textbf{9.53} & \textbf{28.15} \\
 \midrule
 \multirow{3}{*}{10} & PoBEV & \cmark & - & \multirow{3}{*}{50} & 70.00 & 32.75 & \textbf{38.07} & 34.43 & 0.80 & 3.33 & \textbf{34.46} & 9.25 & 27.89 \\
 & \net & \xmark & \xmark & & 70.44 & 33.88 & 33.74 & 41.66 & 3.47 & 3.83 & 31.54 & 9.14 & 28.46 \\
 & \net & \xmark & \cmark & & \textbf{72.40} & \textbf{37.06} & 36.89 & \textbf{43.67} & \textbf{3.90} & \textbf{4.20} & 31.05 & \textbf{9.86} & \textbf{29.88} \\
\midrule
 \multirow{3}{*}{50} & PoBEV & \cmark & - & \multirow{3}{*}{30} & \textbf{72.09} & 35.64 & 36.64 & 42.41 & 1.61 & 3.92 & \textbf{41.41} & 9.77 & 30.44 \\
 & \net & \xmark & \xmark & & 71.93 & 33.59 & 36.43 & 42.63 & 4.05 & 4.30 & 31.44 & \textbf{12.76} & 29.64 \\
 & \net & \xmark & \cmark & & 71.85 & \textbf{37.43} & \textbf{38.76} & \textbf{44.15} & \textbf{5.07} & \textbf{4.54} & 31.07 & 11.54 & \textbf{30.55} \\
\midrule
 \multirow{3}{*}{100} & PoBEV & \cmark & - & \multirow{3}{*}{20} & 70.14 & 35.23 & 34.69 & 40.71 & 2.85 & \textbf{5.63} & \textbf{39.78} & \textbf{14.38} & 30.43 \\
 & \net & \xmark & \xmark & & 71.00 & 36.38 & \textbf{37.76} & 44.13 & 4.47 & 4.37 & 30.98 & 12.76 & 30.23 \\
 & \net & \xmark & \cmark & & \textbf{71.39} & \textbf{37.62} & 37.48 & \textbf{44.38} & \textbf{4.73} & 4.72 & 32.73 & 10.84 & \textbf{30.49} \\
 \bottomrule
 \end{tabular}

\end{table*}

\begin{table*}
\centering
 \caption{Ablation study on the efficacy of various constituent components of Explicit Supervision. All results are reported on the KITTI-360 dataset.}
 \vspace{-0.2cm}
\label{tab:ablation-explicit}
\scriptsize
\setlength\tabcolsep{3.7pt}
 \begin{tabular}{cccc|cccccccc|c}
 \toprule
 \textbf{Accumulation} & \textbf{Depth} & \textbf{Clustering} & \textbf{BBox} & \textbf{Road} & \textbf{Sidewalk} & \textbf{Building} & \textbf{Terrain} & \textbf{Person} & \textbf{2-Wheeler} & \textbf{Car} & \textbf{Truck} & \textbf{mIoU} \\
 \midrule
  \cmark & \cmark & \cmark & \cmark & 71.39 & \textbf{37.62} & 37.48 & 44.38 & \textbf{4.73} & \textbf{4.72} & \textbf{32.73} & 10.84 & \textbf{30.49} \\
  \xmark & \cmark & \cmark & \cmark & 60.27 & 26.75 & 19.89 & 41.22 & 0.44 & 1.43 & 26.06 & 10.01 & 23.26 \\
 \cmark & \xmark & \cmark & \cmark & 66.28 & 33.43 & 27.24 & 38.95 & 4.62 & 3.90 & 21.04 & 9.08 & 25.57 \\
 \cmark & \cmark & \xmark & \xmark &  72.68 & 37.32 & \textbf{37.60} & \textbf{44.91} & 2.17 & 2.49 & 29.96 & \textbf{11.17} & 29.79 \\
 \cmark & \cmark & \cmark & \xmark & \textbf{72.73} & 37.42 & 37.30 & 44.54 & 0.00 & 0.00 & 25.95 & 9.25 & 28.40  \\
 \bottomrule
 \end{tabular}
 \vspace{-0.3cm}
\end{table*}

In this section, we study the impact of various components of our self-supervised pipeline using an ablation study on the KITTI-360 dataset. To this end, we perform two ablative experiments to independently analyze the impact of implicit and explicit supervision on the overall performance.

{\parskip=5pt
\noindent\textbf{Implicit Supervision}: In this experiment, we quantify the impact of implicit supervision on model performance by comparing the IoU metrics obtained when our model \textit{with} and \textit{without} implicit supervision is trained on different percentages of BEV pseudolabels. We thus define five percentage splits, i.e., $0.1\%$, $1\%$, $10\%$, $50\%$, and $100\%$ of BEV pseudolabels, and accordingly sample a fixed subset of images given the split percentage. We ensure equal representation of all scenes in the dataset by independently sampling the given percentage of images from each of the scenes. We also ensure model convergence for all percentage splits by increasing the number of epochs for splits having a lower percentage of BEV pseudolabels. Lastly, we also train the state-of-the-art fully-supervised approach, PoBEV, on the same percentage splits of BEV ground truth labels to act as a benchmark for evaluating the performance of our approach. \cref{tab:ablation-implicit-percentages} presents the results for all the percentage splits.}

We observe that our model with implicit supervision (\net) significantly outperforms both PoBEV as well as our model without implicit supervision (\net*) by more than \SI{7}{pp} for extremely low percentage splits of $0.1\%$ and $1\%$. At such low sample counts, PoBEV suffers from the lack of BEV ground truth data while \net~is able to leverage FV training to generate good results in BEV. A large part of the gain can be attributed to the better segmentation of static classes which is a direct consequence of the robust supervision generated from the warping step of implicit training. We also observe that \net* outperforms PoBEV by more than \SI{1.5}{pp} at these percentages which highlights the training efficiency of our network architecture. Further, we observe that with only $10\%$ of the labels, \net~almost matches the state-of-the-art result, while \net* and PoBEV are \SI{2.09}{pp} and \SI{2.66}{pp} away respectively. This observation further emphasizes the benefit of incorporating implicit supervision into the training procedure to circumvent the need for expensive BEV ground truth annotations. However, from this percentage split onwards, we observe that PoBEV starts outperforming \net~for the \textit{car} class which can be attributed to the better and consistent supervision of BEV ground truth labels. As the percentage of BEV samples increases, all approaches converge to a similar mIoU score since the labels in BEV compensate for the privileged information supplied by implicit supervision. However, we still note that \net~outperforms PoBEV for all static classes which demonstrates the ability of implicit supervision to positively augment the training procedure even in the presence of $100\%$ of BEV labels.
To further demonstrate the impact of implicit supervision on the trained model, we provide qualitative results for every percentage split in the supplementary material.

{\parskip=5pt
\noindent\textbf{Explicit Supervision}: In this experiment, we study the influence of various components of our BEV pseudolabel generation pipeline by removing each component from the overall solution proposed in \cref{subsec:explicit-supervision} and analyzing the change in the overall model performance. \cref{tab:ablation-explicit} outlines the results of this ablation study.
The first row of \cref{tab:ablation-explicit} shows the results of the model trained with all components of our pseudolabel generation pipeline and acts as a baseline for evaluating the impact of each constituent module. 
We observe in the second row that frame accumulation forms the most important step of the pseudolabel generation pipeline whose removal results in a drop of \SI{7.23}{pp}. Frame accumulation improves the density of static regions and helps the model reason about far-away and occluded regions. In the third row, we replace the depth-based lifting of static classes for pseudolabel creation with the IPM algorithm and observe a drop in performance of both static and dynamic classes. We attribute this behavior to the strong correlation between static and dynamic classes as they are predicted in a joint manner within a single map. The last two rows indicate that the outlier removal and incorporation of prior knowledge on dynamic classes yield a large performance gain w.r.t. cars (\SI{7.23}{pp} and \SI{6.78}{pp} respectively) but have a minor impact on static classes and trucks.}

\begin{figure*}
\centering
\footnotesize
\setlength{\tabcolsep}{0.05cm}% for the horiz padding
{
\renewcommand{\arraystretch}{0.2}% for the vertical padding
\newcolumntype{M}[1]{>{\centering\arraybackslash}m{#1}}
\begin{tabular}{cM{5.6cm}M{2.8cm}M{2.8cm}M{2.8cm}}
& Input FV Image & PanopticBEV~\cite{cit:bev-seg-panopticbev} & \net~(Ours) & Improvement/Error Map \\
\\
(a) & {\includegraphics[width=\linewidth, height=0.455\linewidth, frame]{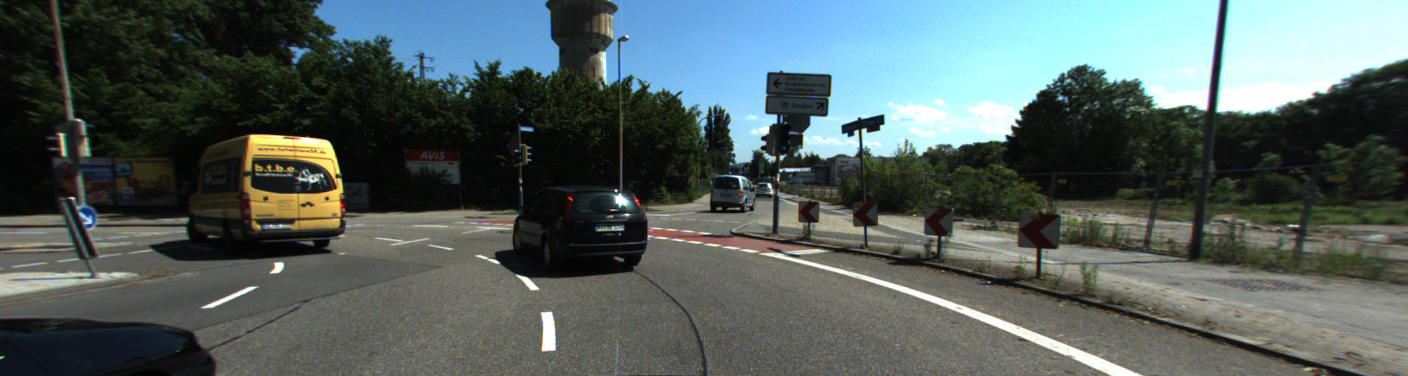}} & {\includegraphics[width=\linewidth, frame]{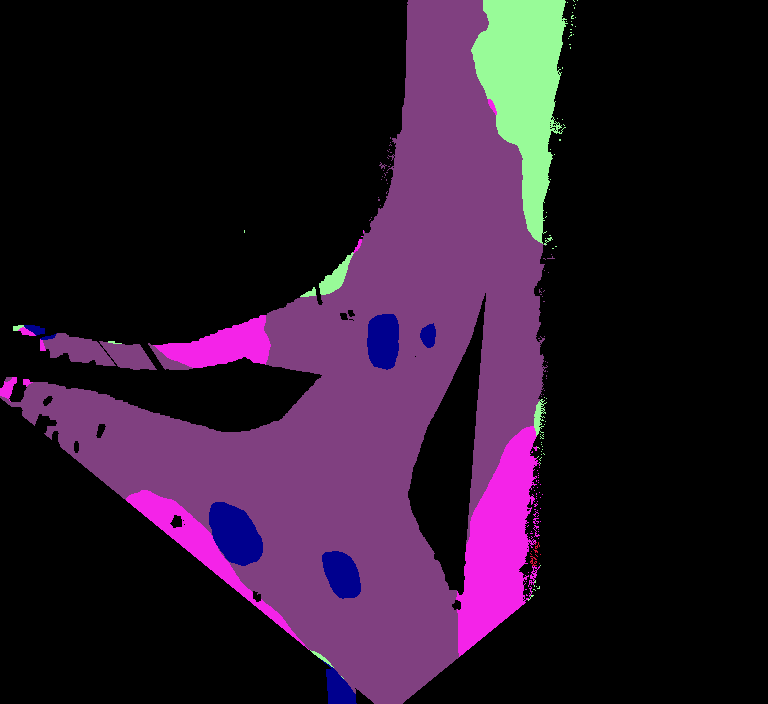}} & {\includegraphics[width=\linewidth, frame]{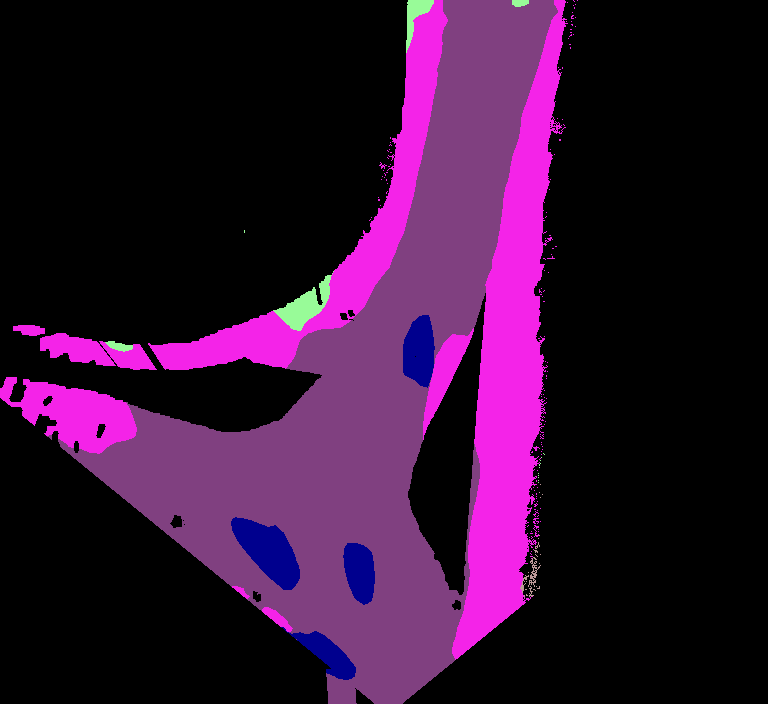}} & {\includegraphics[width=\linewidth, frame]{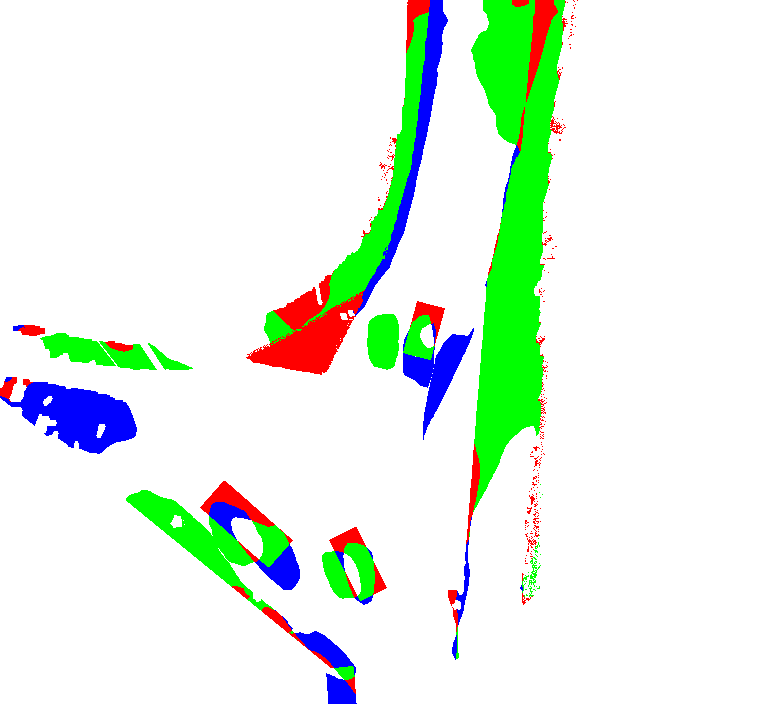}} \\
\\
(b) & \includegraphics[width=\linewidth, height=0.455\linewidth, frame]{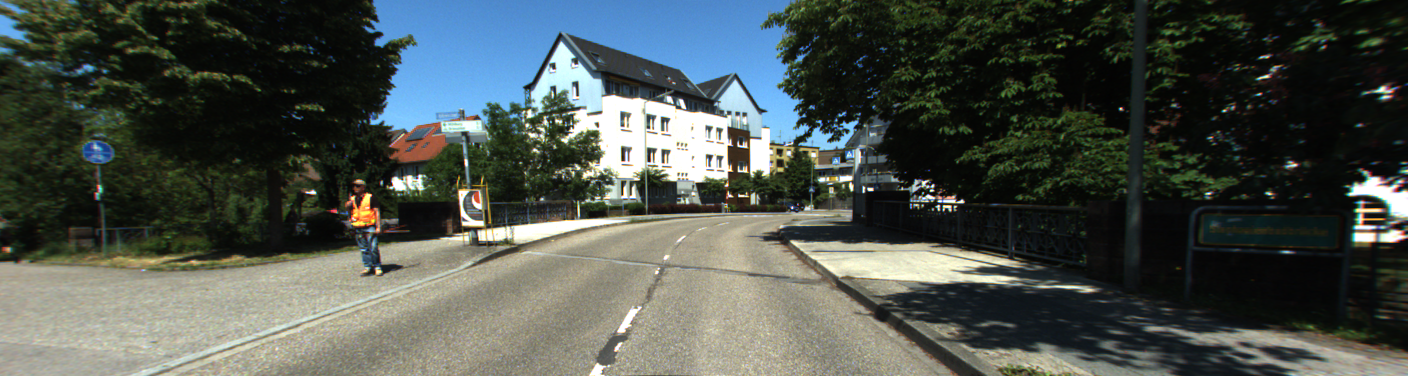} & \includegraphics[width=\linewidth, frame]{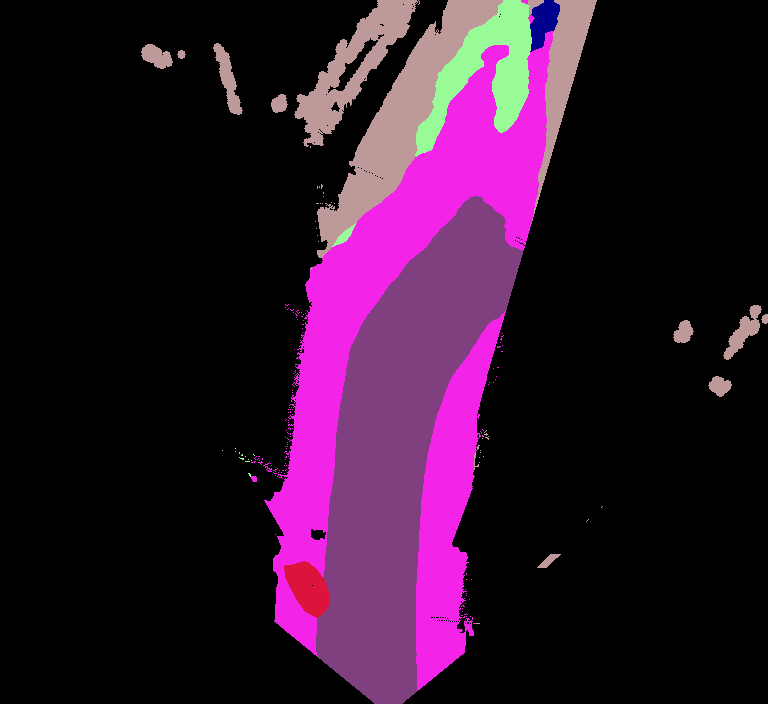} & \includegraphics[width=\linewidth, frame]{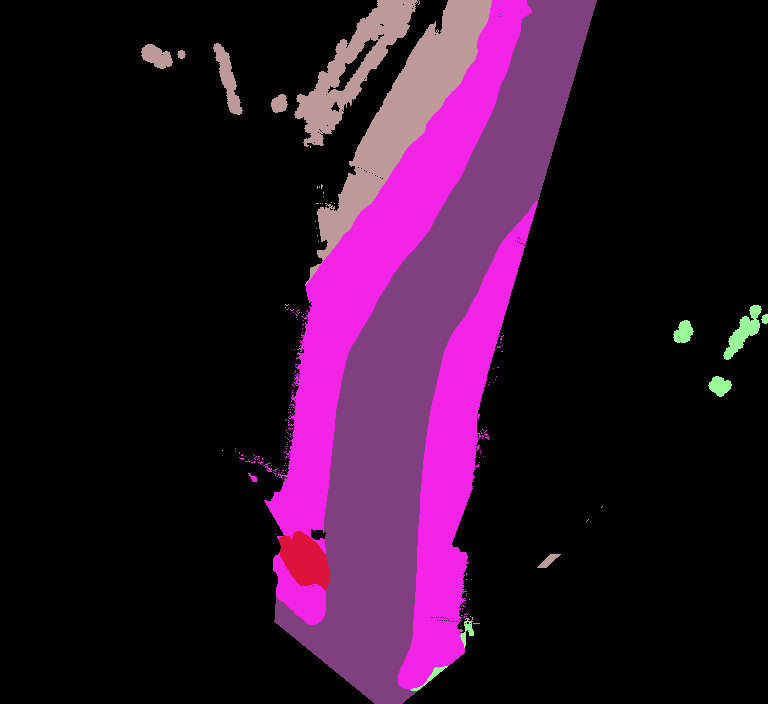} & \includegraphics[width=\linewidth, frame]{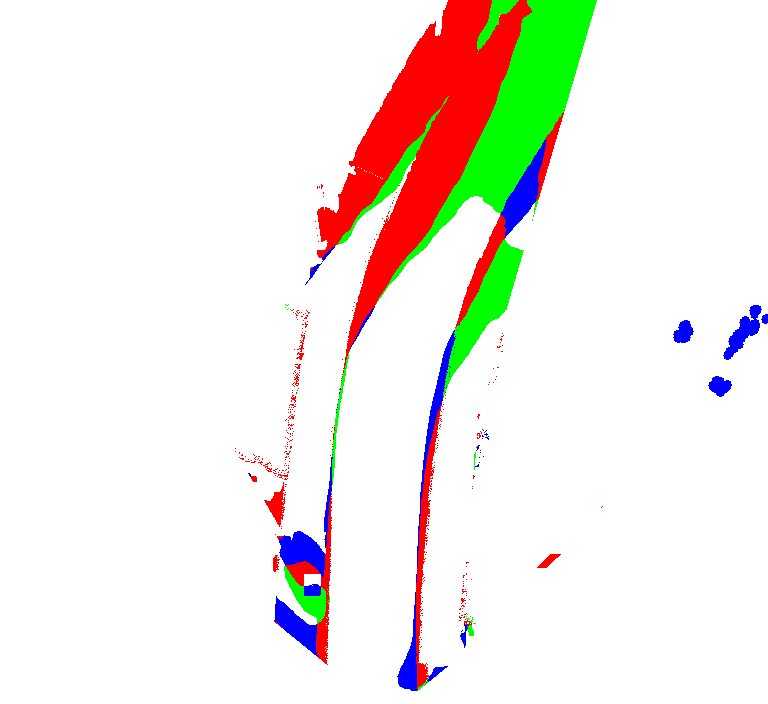} \\
\\
(c) & \includegraphics[width=\linewidth, height=0.455\linewidth, frame]{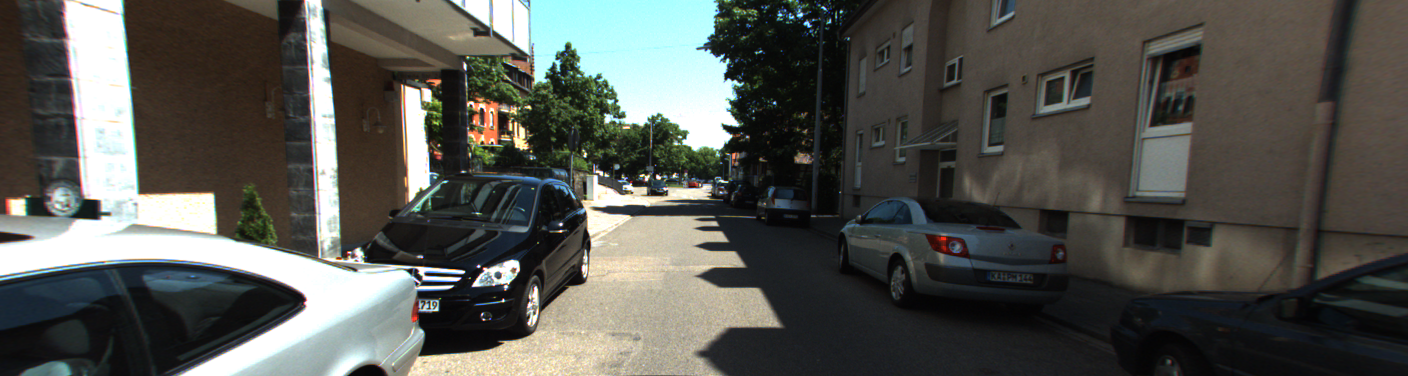} & \includegraphics[width=\linewidth, frame]{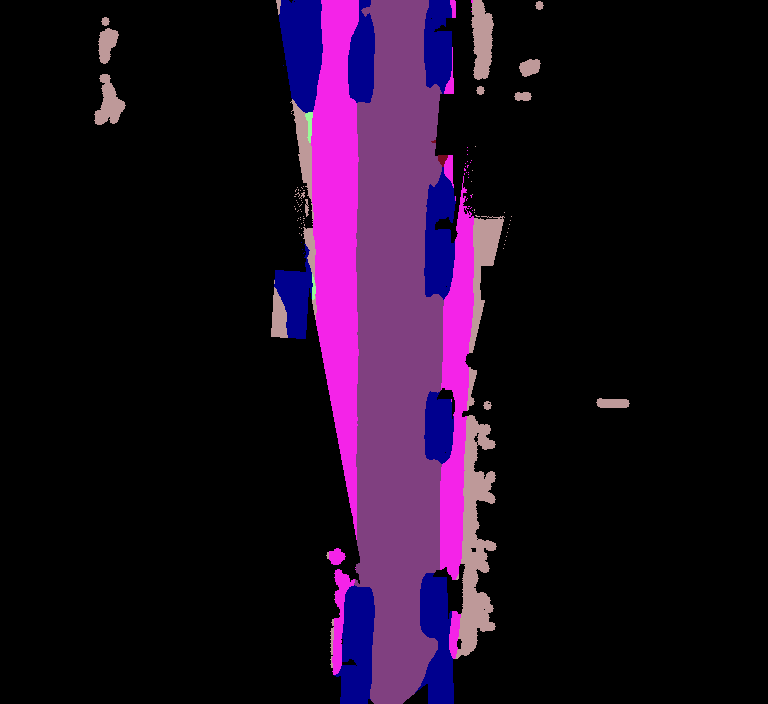} & \includegraphics[width=\linewidth, frame]{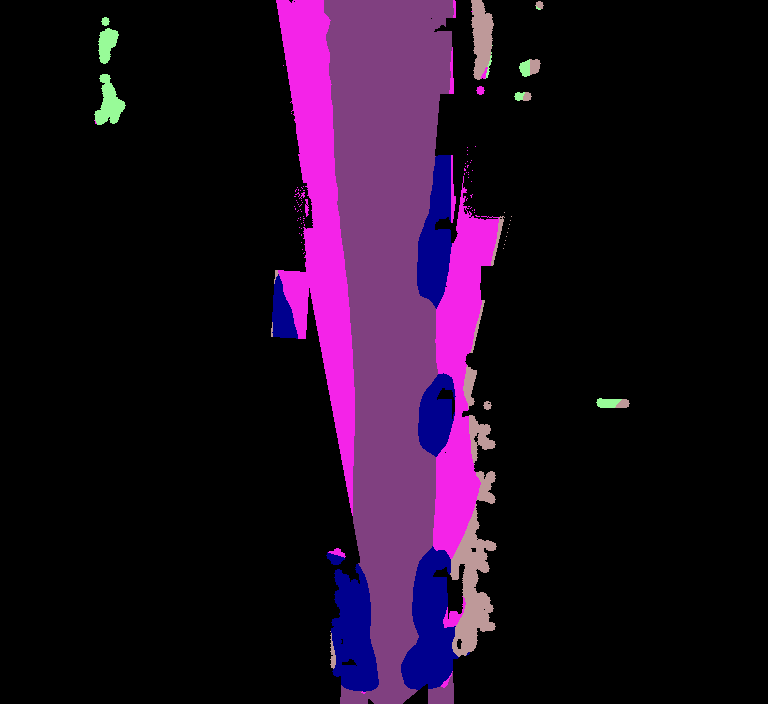} & \includegraphics[width=\linewidth, frame]{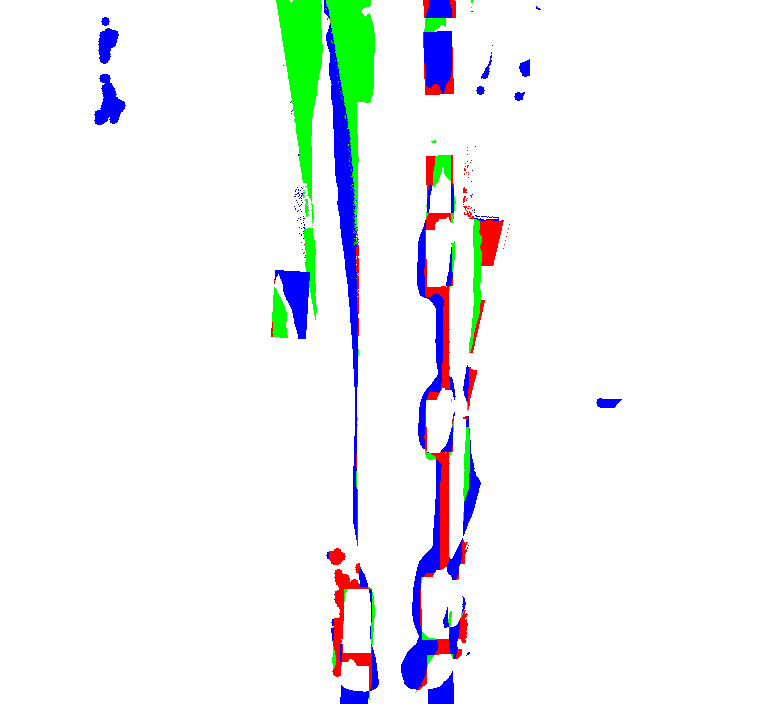} \\
\\
(d) & \includegraphics[width=\linewidth, height=0.455\linewidth, frame]{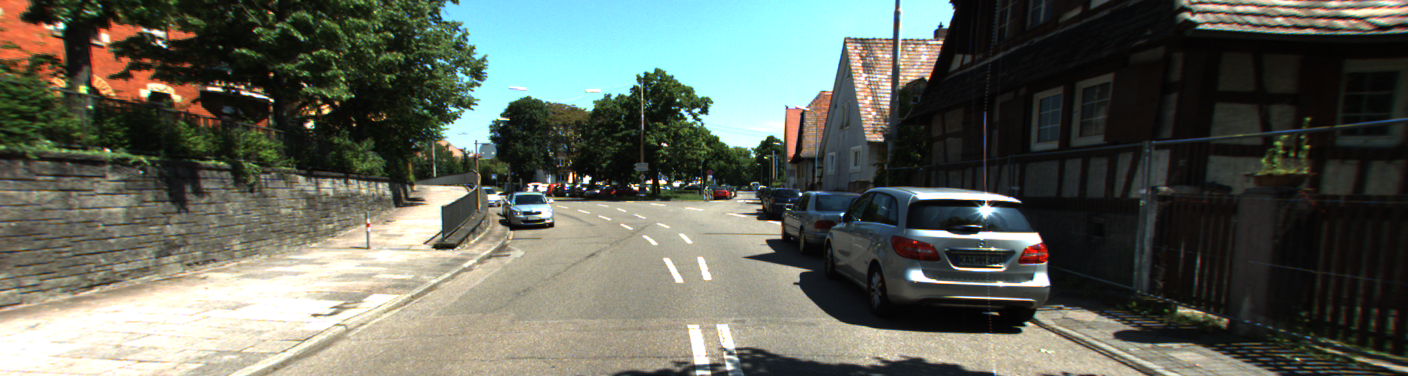} & \includegraphics[width=\linewidth, frame]{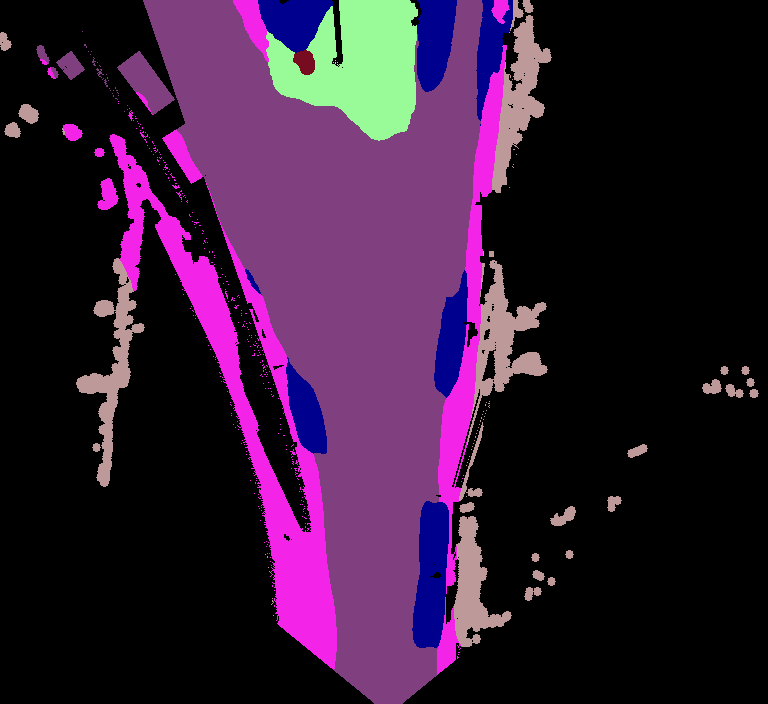} & \includegraphics[width=\linewidth, frame]{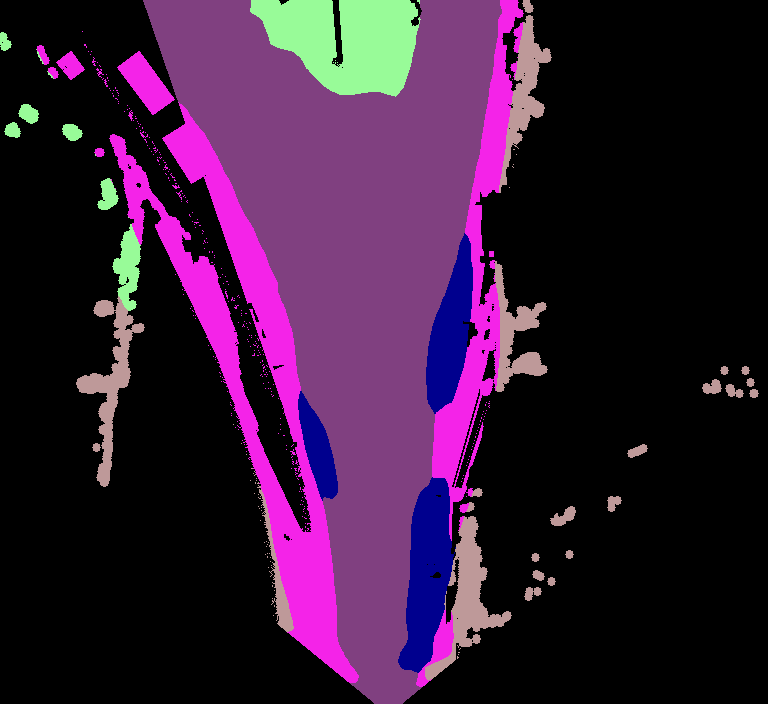} & \includegraphics[width=\linewidth, frame]{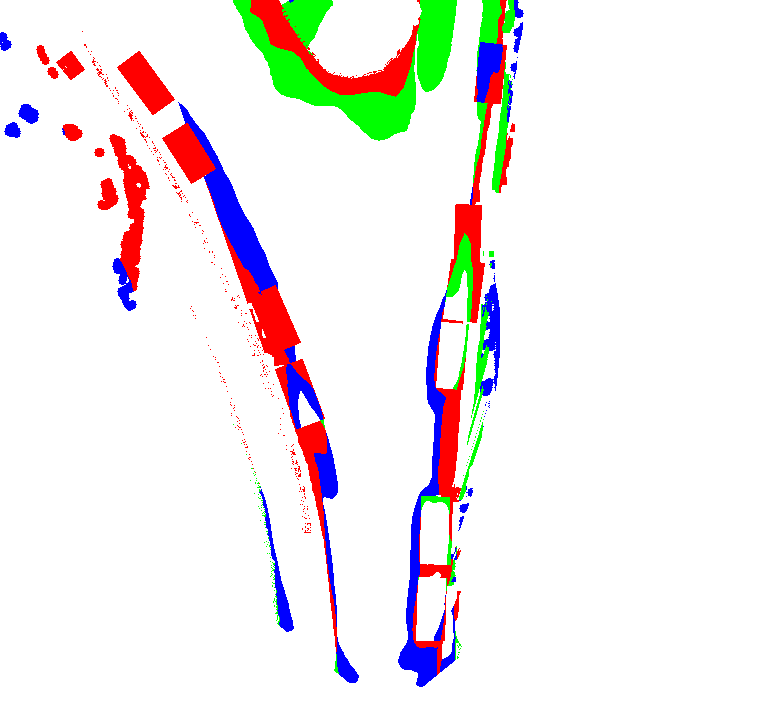} \\
\end{tabular}
}
\caption{Qualitative results of our self-supervised framework \net~in comparison with PanopticBEV~\cite{cit:bev-seg-panopticbev} on the KITTI-360 dataset. We also show the Improvement/Error map which shows pixels misclassified by PanopticBEV and correctly predicted by \net~in green, pixels misclassified by \net~and correctly by PanopticBEV in blue, and pixels misclassified by both models in red.}
\label{fig:qual-analysis}
\vspace{-0.3cm}
\end{figure*}

%%%%%%%%%%%%%%%%%%%%%%%%%%%%%%%%%%%%%%%%%%%% GENERALIZATION EXPERIMENTS %%%%%%%%%%%%%%%%%%%%%%%%%%%%%%%%%%%%%%%
\subsection{Generalization Experiments}
\label{subsec:generalization-experiments}

% Generalization Experiments
\begin{table}
\centering
\caption{Evaluation of model generalizability across datasets. Models pre-trained on KITTI-360 are evaluated on Waymo.}
\vspace{-0.2cm}
\label{tab:generalizability-kitti2waymo}
\scriptsize
\setlength\tabcolsep{3.7pt}
\begin{tabular}{c|ccccc|c}
\toprule
\textbf{Method} & TIIM & VED & VPN & PON & PoBEV & \net~(Ours) \\
\midrule
\textbf{mIoU} & 16.53 & 14.02 & 13.52 & 12.05 & 16.94 & \textbf{22.57} \\
\bottomrule
\end{tabular}
\vspace{-0.3cm}
\end{table}

We evaluate the generalization ability of \net~by evaluating the best model from the KITTI-360 dataset on the Waymo dataset and comparing its performance to that of the other baselines. VED, VPN, PON, and PoBEV use image dimensions as channel count in their learnable layers which forces the Waymo image size to be equal to that of KITTI-360 ($1408\times384$). TIIM and \net~are agnostic to the image size and we thus report the result obtained when using images of size $512\times352$ which aligns the Waymo camera intrinsics with that of KITTI-360. \cref{tab:generalizability-kitti2waymo} presents the results of this experiment. We observe that our approach generalizes significantly better to the Waymo dataset as compared to other baselines outperforming them by \SI{5.63}{pp}. This large gain in mIoU can be credited to the superior geometric reasoning and world modeling of our approach which is largely facilitated by our self-supervised learning framework. Our implicit supervision guides the network to learn geometrically coherent features which helps it in generalizing well across different datasets. On the contrary, the BEV ground truths do not enforce geometric consistency which results in the poor performance of the fully supervised baselines.

%%%%%%%%%%%%%%%%%%%%%%%%%%%%%%%%%% QUALITATIVE EVALUATION %%%%%%%%%%%%%%%%%%%%%%%%%%%%%%%%%%
\subsection{Qualitative Results}

We further evaluate \net~by qualitatively comparing it with the state-of-the-art fully supervised approach, PoBEV, in \cref{fig:qual-analysis}. We observe from \cref{fig:qual-analysis}(a) that both PoBEV and \net~are able to capture the characteristics of close regions and are also able to localize nearby vehicles to a high accuracy. Further, as evident from the error/improvement maps in the last column, our approach precisely captures the contour of the sidewalk for near as well as distant regions. A similar observation can be made in \cref{fig:qual-analysis}(b) where our model accurately predicts the curve of the road over long distances, but PoBEV fails to do so and instead confuses road with sidewalk. This superiority in inferring static elements of the scene can be attributed to implicit supervision which encourages the network to learn consistent features over long horizons. 
\cref{fig:qual-analysis}(c, d) demonstrate that our network is able to successfully localize a large number of vehicles in the scene. These images also depict a limitation of our network, i.e., the failure to predict cars in distant regions. This behavior can be attributed to the extreme sparsity of dynamic objects in distant regions which results in the generation of sub-optimal BEV pseudolabels. This limitation, however, is not faced by PoBEV which uses BEV ground truth labels and thus receives reliable supervision throughout the BEV image. We also observe that our network accurately estimates the contours of \textit{road}, \textit{sidewalk}, and \textit{terrain}, thus highlighting the benefit of implicit supervision. We provide further qualitative results of our approach in the supplementary material.

\subsection{Discussion of Limitations}
\label{subsec:limitations}
While our self-supervised approach, \net, performs on par with fully supervised state-of-the-art approaches, it is subject to two kinds of limitations. Firstly, the reliance on temporal context can deteriorate its performance in highly dynamic scenes where moving objects may produce artifacts in both the generated pseudolabels and the implicit supervision signal. Here, explicit handling of moving objects can help minimize these effects. Secondly, perspective distortion limits the spatial observability of the scene for distant regions. This, however, is a long-standing limitation of camera-based methods that can be overcome using a longer temporal baseline in conjunction with a dedicated dynamic object handling strategy. 
\section{Conclusion}
\label{sec:conclusion}
In this paper, we present \net, the first self-supervised approach to generate a semantic map in BEV using a single FV monocular image, thus alleviating the need for expensive BEV semantic ground truth annotations. Our approach relies on only FV image sequences and their corresponding FV semantic annotations to generate two modes of supervision, namely, implicit supervision and explicit supervision. Using extensive evaluations on the KITTI-360 dataset, we demonstrate that \net~performs on par with the state-of-the-art fully supervised BEV approaches while already achieving competitive performance with only $1\%$ of pseudolabels in the BEV. Future work includes relaxing the requirement for FV semantic ground truth labels and instead relying on coarse FV predictions from a generic pre-trained FV semantic network or leveraging scene knowledge from unsupervised FV semantic segmentation approaches.

{\parskip=3pt
\noindent\textbf{Acknowledgements}: This work was funded by the German Research Foundation (DFG) Emmy Noether Program grant number 468878300 and a hardware grant from NVIDIA.}

%%%%%%%%% REFERENCES
{\small
\bibliographystyle{ieee_fullname}
\bibliography{references}
}

\flushcolsend

%%% Add the supplementary %%%

%%%%%%%%%% Merge with supplemental materials %%%%%%%%%%
\clearpage

\begin{strip}
\begin{center}
\vspace{-5ex}
\textbf{\Large \bf
SkyEye: Self-Supervised Bird's-Eye-View Semantic Mapping\\Using Monocular Frontal View Images
} \\
\vspace{2ex}

\Large{\bf- Supplementary Material -}\\
\vspace{0.4cm}
\normalsize{Nikhil Gosala$^{*}$, K\"ursat Petek$^{*}$, Paulo L. J. Drews-Jr, Wolfram Burgard, and Abhinav Valada}
\end{center}
\end{strip}

%%%%%%%%%% Merge with supplemental materials %%%%%%%%%%
%%%%%%%%%% Prefix a "S" to all equations, figures, tables and reset the counter %%%%%%%%%%
\setcounter{section}{0}
\setcounter{equation}{0}
\setcounter{figure}{0}
\setcounter{table}{0}
\makeatletter

\renewcommand{\thesection}{S.\arabic{section}}
\renewcommand{\thesubsection}{S.\arabic{section}.\arabic{subsection}}
\renewcommand{\thetable}{S.\arabic{table}}
\renewcommand{\thefigure}{S.\arabic{figure}}

\normalsize

In this supplementary material, we present additional experiments to study the performance of \net. Specifically, we analyze the computational efficiency of our model in \cref{sec:supp-model-efficiency} and perform an additional ablative experiment to study the characteristics of our contribution in \cref{sec:supp-ablation-experiments}. We also present further qualitative results to demonstrate the performance of \net~on the KITTI-360 dataset in \cref{sec:supp-qualitative-results}.

\section{Evaluation of Model Efficiency}
\label{sec:supp-model-efficiency}

In this section, we compare the efficiency of \net~with that of the other baselines on the KITTI-360 dataset. \cref{tab:supp-network-efficiency} presents the results of this comparison on an nVidia GTX 3090 GPU. We observe that \net~is the most efficient of all baselines in terms of the number of learnable parameters, using $7.73$ million fewer parameters as compared to the state-of-the-art fully supervised approach, PoBEV~\cite{cit:bev-seg-panopticbev}. We also note that \net~uses significantly fewer Multiply-Accumulate (MAC) operations as compared to PoBEV. A large part of this efficiency can be attributed to the use of a 3D voxel grid representation to jointly represent features of both FV and BEV, thus alleviating the need for separate representations in FV and BEV. Further, we observe that our network design translates into a nearly $3$-fold reduction in runtime as compared to PoBEV, requiring only \SI{77.84}{\milli \second} per sample and allowing \net~to be used in many real-time applications. 
 
 % Kitti-360 Efficiency
\begin{table}[htb]
\footnotesize
\centering
\caption{Comparison of model efficiency on the KITTI-360 dataset.}
\label{tab:supp-network-efficiency}
\setlength\tabcolsep{3.7pt}
 \begin{tabular}{c|ccc}
 \toprule
\textbf{Method} & \textbf{\# Params (M)} & \textbf{MAC (G)} & \textbf{Runtime (\SI{}{\milli\second})} \\
 \midrule
IPM~\cite{cit:bev-seg-ipm-original} & 27.83 & \textbf{32.51} & 71.65 \\
VED~\cite{cit:bev-seg-lu2019ved} & 169.20 & 391.49 & 25.42 \\
VPN~\cite{cit:bev-seg-pan2020vpn} & 23.01 & 55.45 & \textbf{8.68} \\
PON~\cite{cit:bev-seg-pon} & 91.06 & 657.63 & 164.08 \\
TIIM~\cite{cit:bev-seg-tiim} & 40.72 & 1290.16 & 193.82 \\
PoBEV~\cite{cit:bev-seg-panopticbev} & 22.33 & 272.68 & 213.92 \\
 \cmidrule{1-4}
Ours & \textbf{14.60} & 219.94 & 77.84 \\
\bottomrule
 \end{tabular}
\vspace{-0.5cm}
\end{table}

\section{Additional Ablation Experiments}
\label{sec:supp-ablation-experiments}
In this section, we perform an ablation experiment to analyze the impact of different FV window sizes (WS) used during implicit supervision on the overall performance of the model. To this end, we perform implicit supervision using windows of size $0$, $6$, $10$, $14$, and $20$, and subsequently refine these pre-trained models using $1\%$ of BEV pseudo labels. We choose $1\%$ of BEV pseudo labels to better highlight the impact of window sizes on the overall performance of the model. \cref{tab:network-ablation-windowsize} presents the results of this ablation study. We observe that a window size of 10 generates the best mIoU score across all the tested window sizes, marginally outperforming a window size of 6 by \SI{0.12}{pp}. We also observe that larger window sizes report worse performance across almost all classes. Thus, we use $\text{WS}=10$ in our \net~framework and perform all experiments with this window size. 

\begin{table*}[htb]
\centering
\footnotesize
\caption{Ablation study on the impact of Window Size on the overall performance of the model. This experiment uses only $1$\% of BEV pseudo labels to highlight the impact of window size. All scores are reported on the KITTI-360 dataset.}
\label{tab:network-ablation-windowsize}
\setlength\tabcolsep{3.7pt}
 \begin{tabular}{c|cccccccc|c}
 \toprule
  \textbf{Window Size} & \textbf{Road} & \textbf{Sidewalk} & \textbf{Building} & \textbf{Terrain} & \textbf{Person} & \textbf{2-Wheeler} & \textbf{Car} & \textbf{Truck} & \textbf{mIoU} \\
 \midrule
 0 & 71.82 & 33.27 & 36.00 & 41.23 & 2.80 & 0.00 & 24.97 & 5.42 & 26.94 \\
 6 & 71.97 & 35.63 & 36.86 & 39.07 & 2.93 & 0.00 & 28.23 & 9.55 & 28.03 \\
 10 & 72.00 & 33.76 & 37.59 & 38.75 & 3.77 & 1.81 & 28.04 & 9.53 & \textbf{28.15} \\
 14 & 71.81 & 32.87 & 36.93 & 38.47 & 3.02 & 2.87 & 28.58 & 7.93 & 27.81 \\
 20 & 71.21 & 35.16 & 35.77 & 38.26 & 3.59 & 2.35 & 28.91 & 5.16 & 27.43 \\
 \bottomrule
 \end{tabular}
\end{table*}

\section{Additional Qualitative Results}
\label{sec:supp-qualitative-results}
In this section, we present additional qualitative results on the KITTI-360 dataset and also present the BEV semantic maps obtained when our \net~model is trained using different percentages of BEV pseudo labels. 

\subsection{BEV Semantic Mapping}

\begin{figure*}
\centering
\footnotesize
\setlength{\tabcolsep}{0.05cm}% for the horiz padding
{
\renewcommand{\arraystretch}{0.2}% for the vertical padding
\newcolumntype{M}[1]{>{\centering\arraybackslash}m{#1}}
\begin{tabular}{cM{5.6cm}M{2.8cm}M{2.8cm}M{2.8cm}}
& Input FV Image & PanopticBEV~\cite{cit:bev-seg-panopticbev} & \net~(Ours) & Improvement/Error Map \\
\\
(a) & {\includegraphics[width=\linewidth, height=0.455\linewidth, frame]{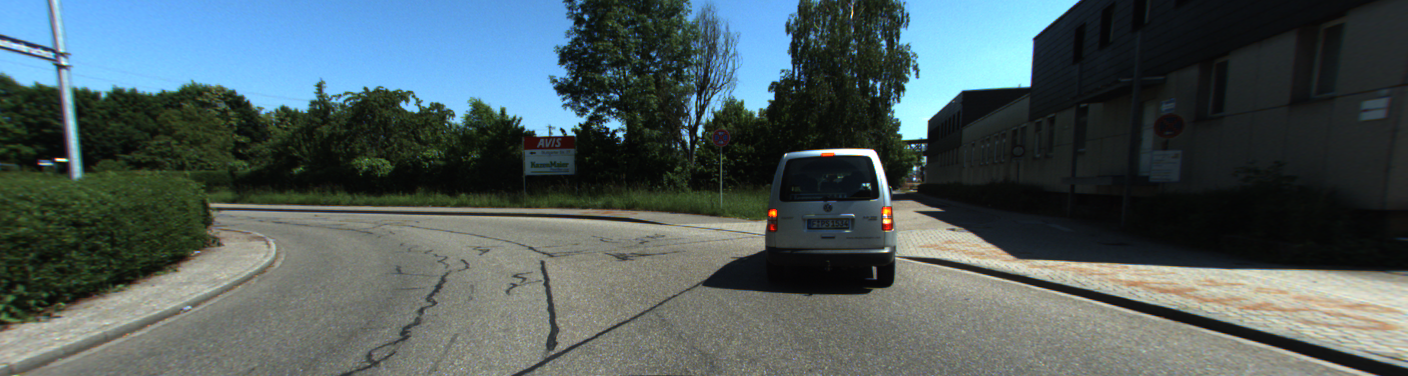}} & {\includegraphics[width=\linewidth, frame]{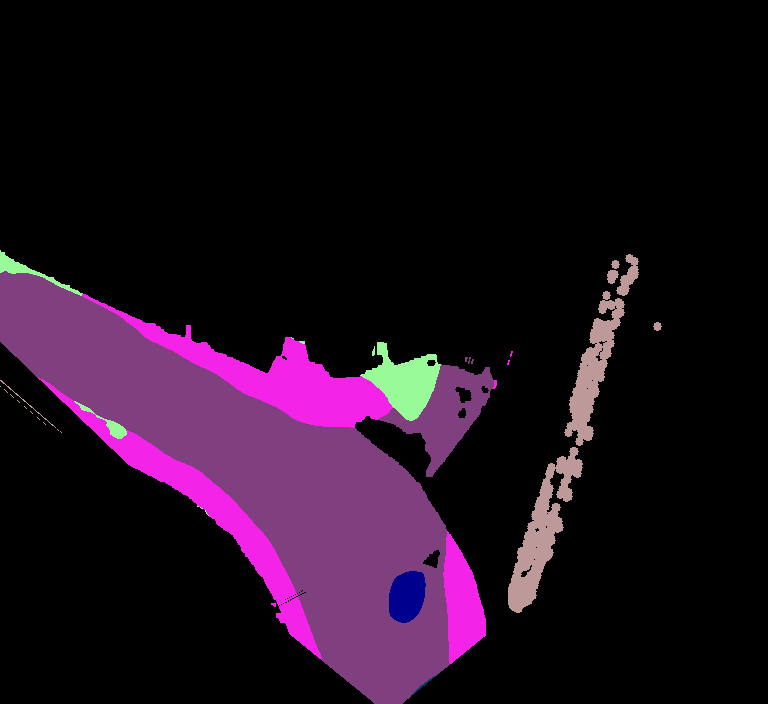}} & {\includegraphics[width=\linewidth, frame]{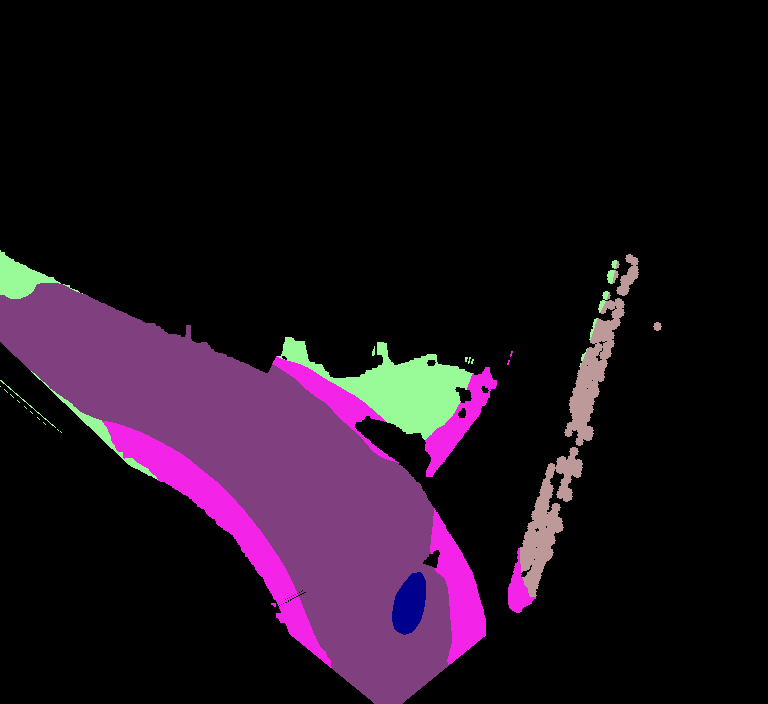}} & {\includegraphics[width=\linewidth, frame]{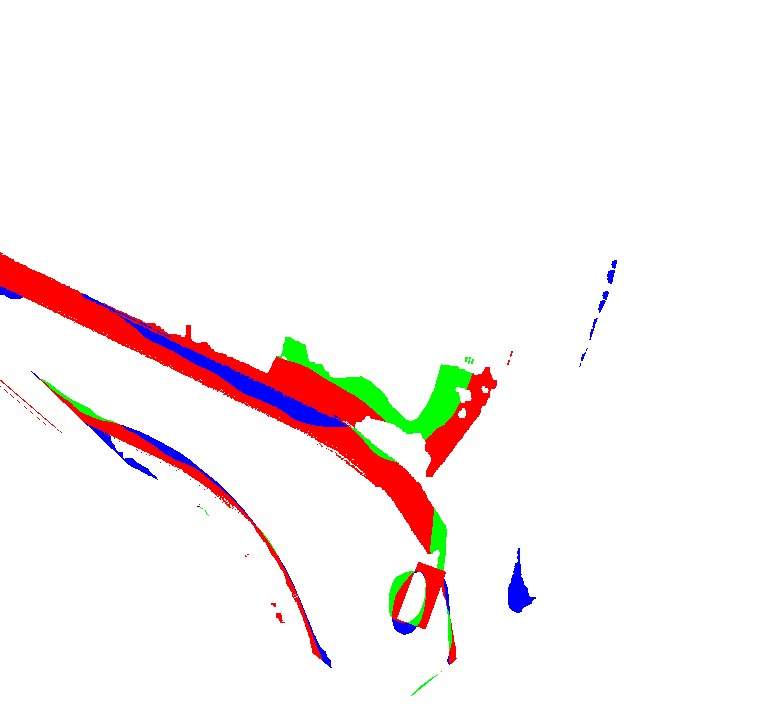}} \\
\\
(b) & \includegraphics[width=\linewidth, height=0.455\linewidth, frame]{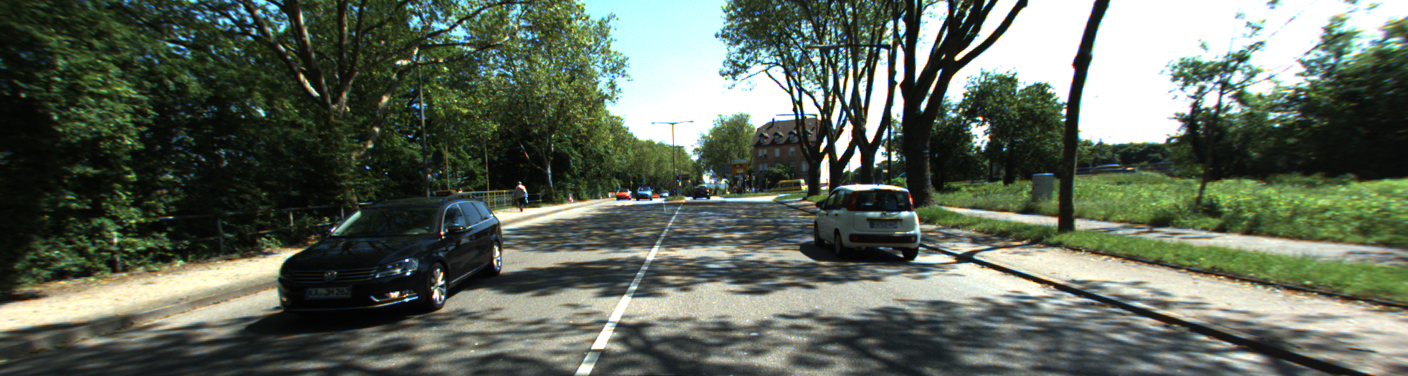} & \includegraphics[width=\linewidth, frame]{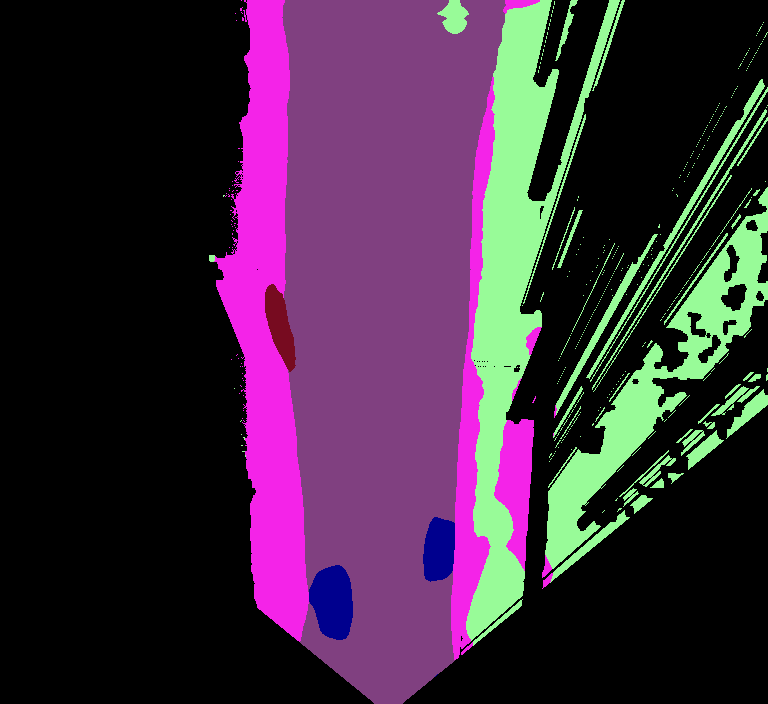} & \includegraphics[width=\linewidth, frame]{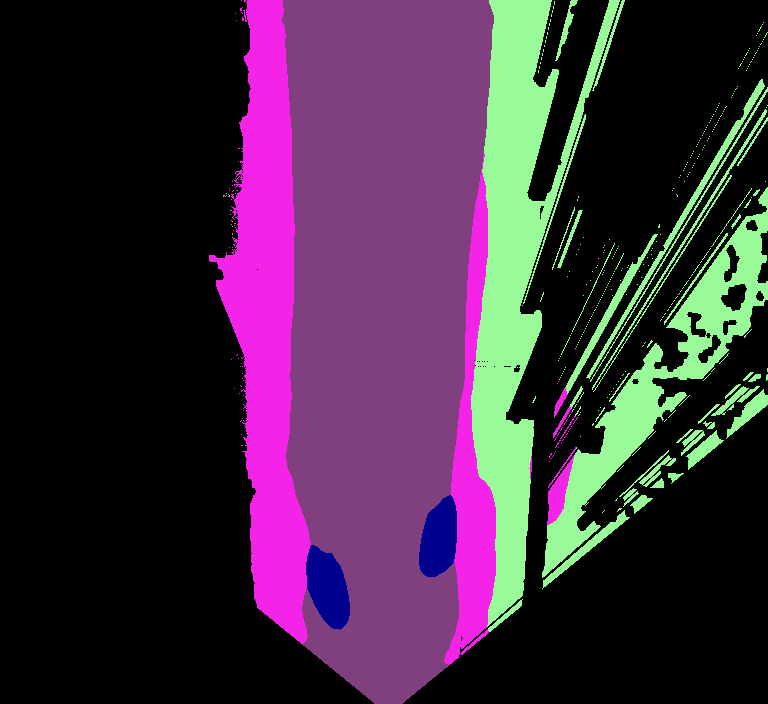} & \includegraphics[width=\linewidth, frame]{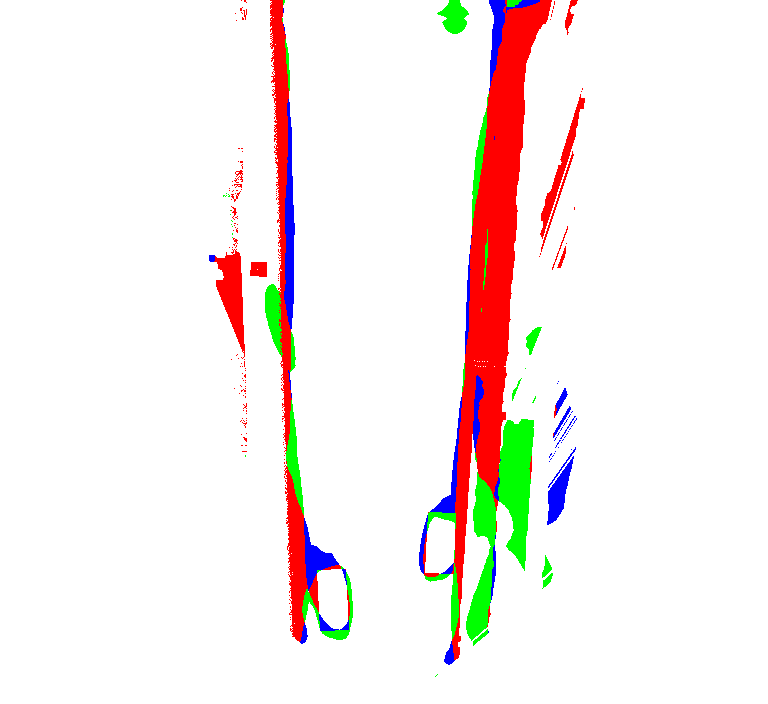} \\
\\
(c) & \includegraphics[width=\linewidth, height=0.455\linewidth, frame]{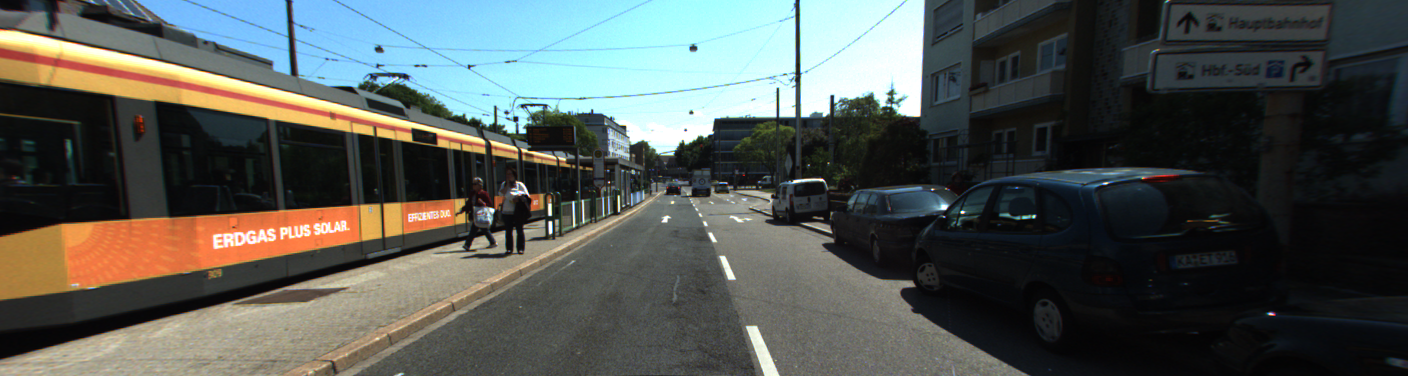} & \includegraphics[width=\linewidth, frame]{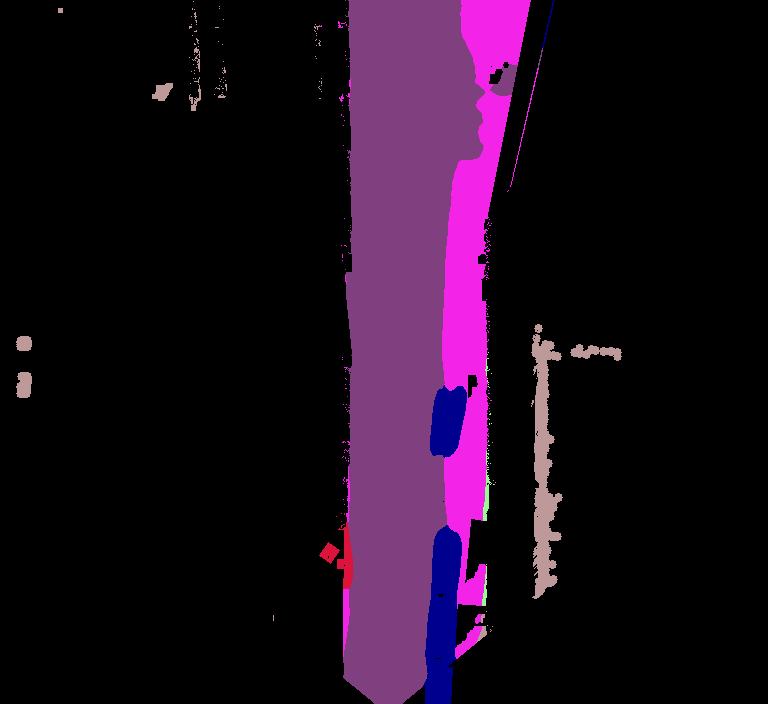} & \includegraphics[width=\linewidth, frame]{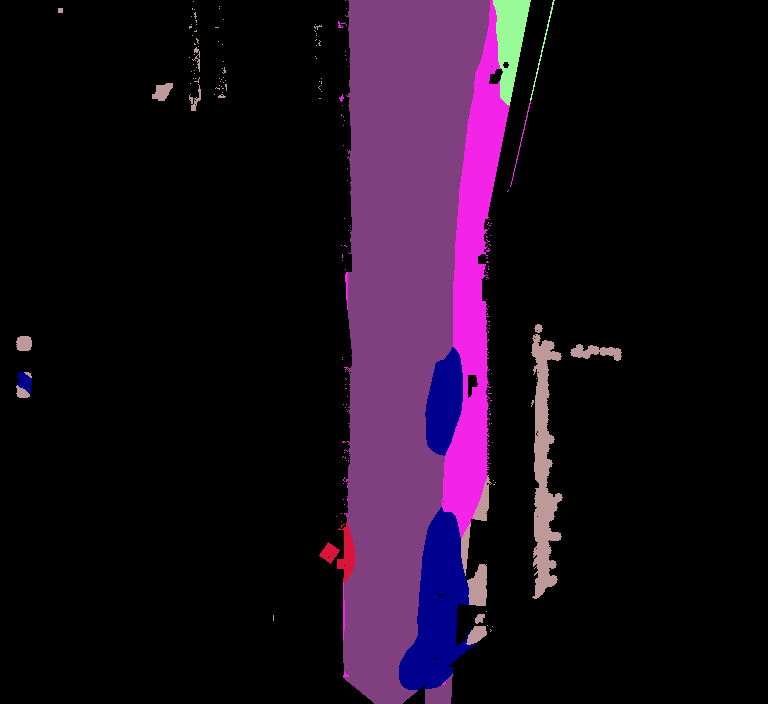} & \includegraphics[width=\linewidth, frame]{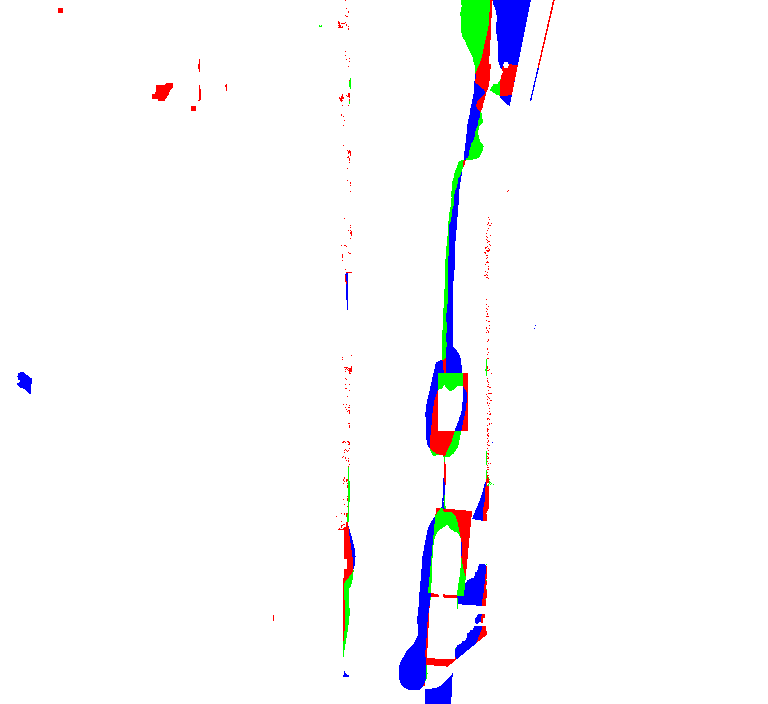} \\
\\
(d) & \includegraphics[width=\linewidth, height=0.455\linewidth, frame]{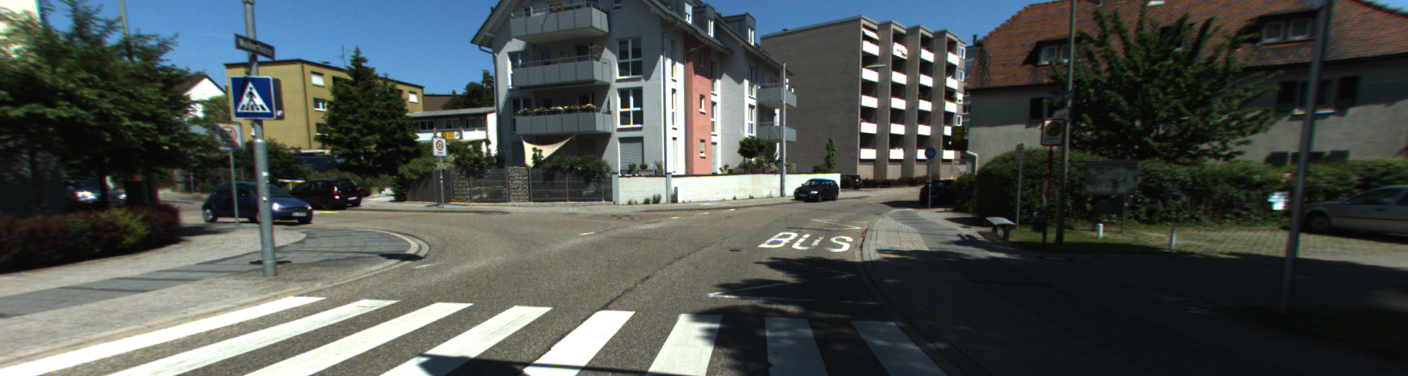} & \includegraphics[width=\linewidth, frame]{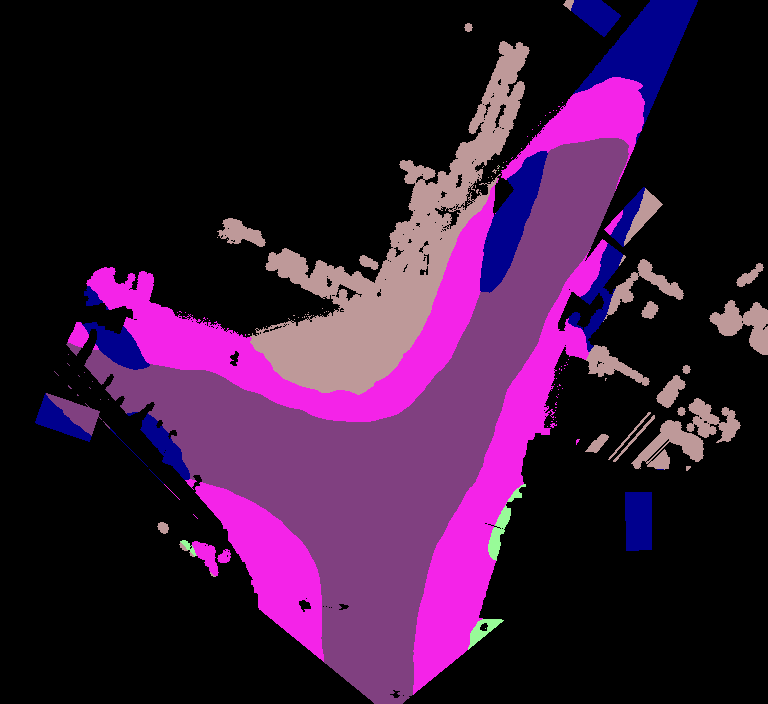} & \includegraphics[width=\linewidth, frame]{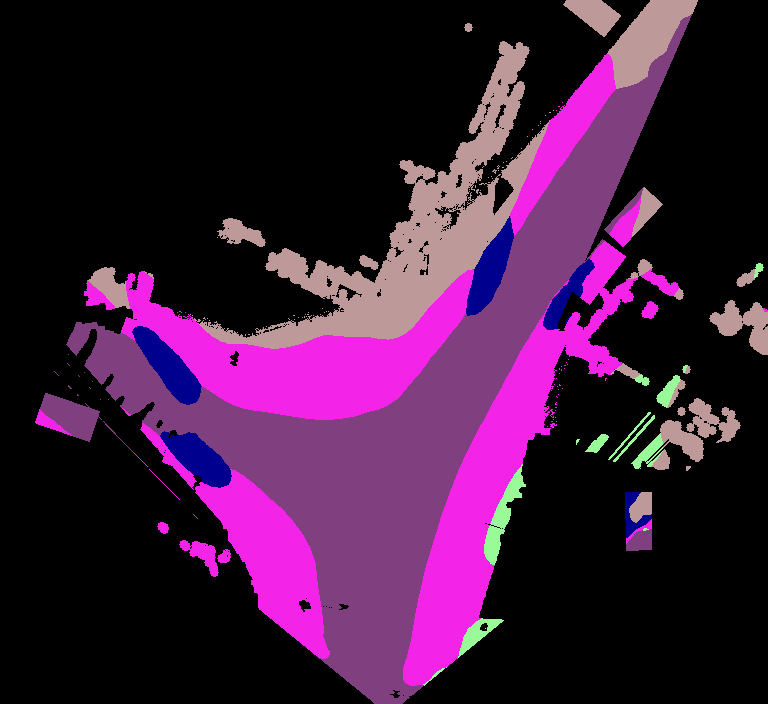} & \includegraphics[width=\linewidth, frame]{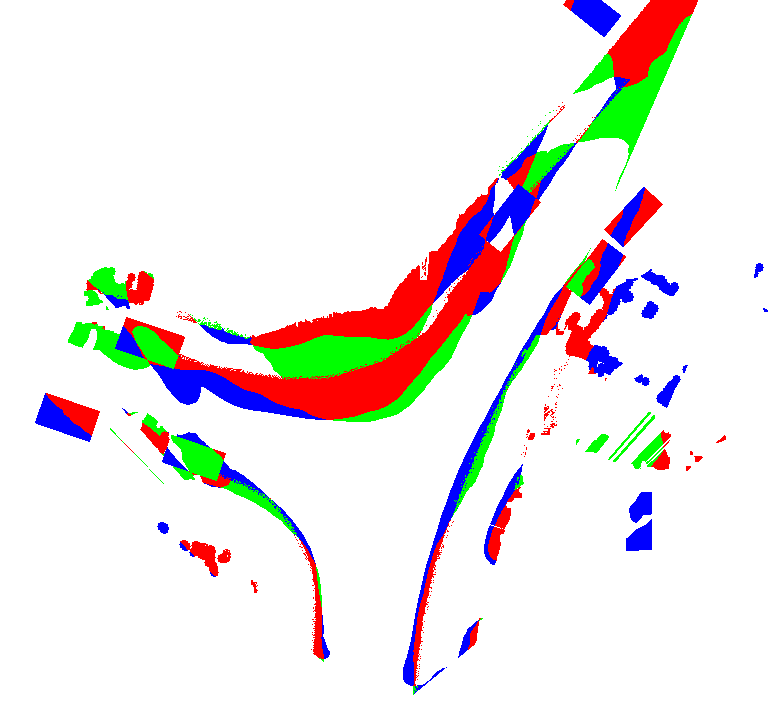} \\
\\
(e) & \includegraphics[width=\linewidth, height=0.455\linewidth, frame]{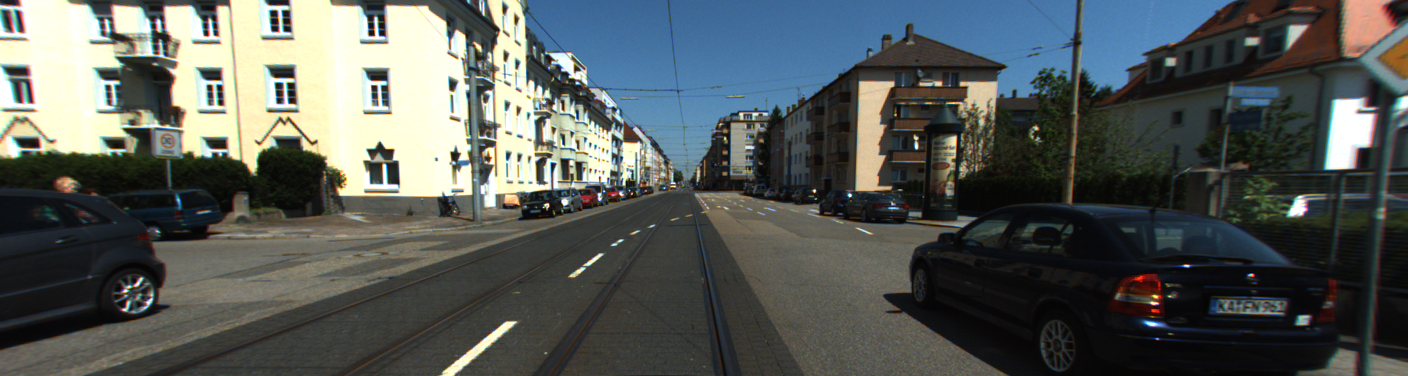} & \includegraphics[width=\linewidth, frame]{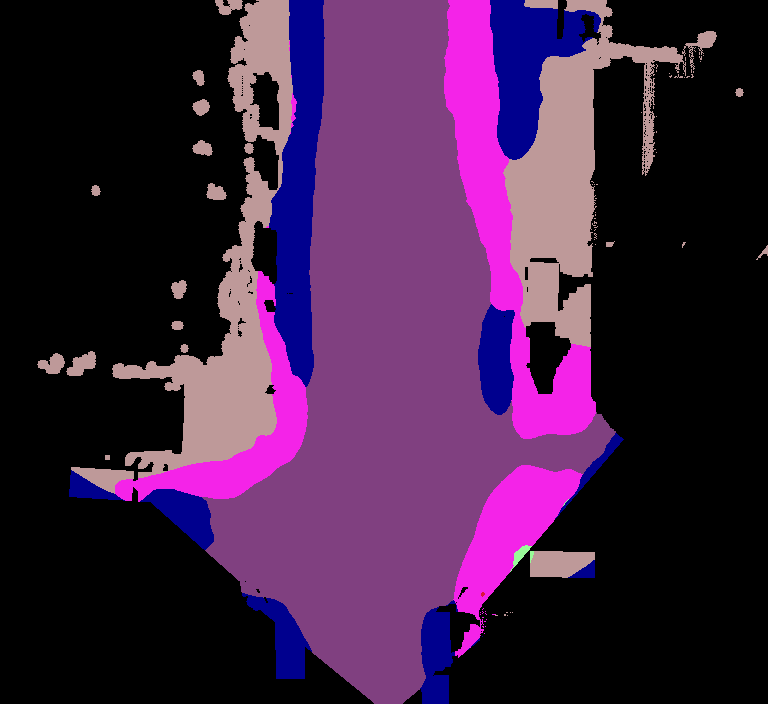} & \includegraphics[width=\linewidth, frame]{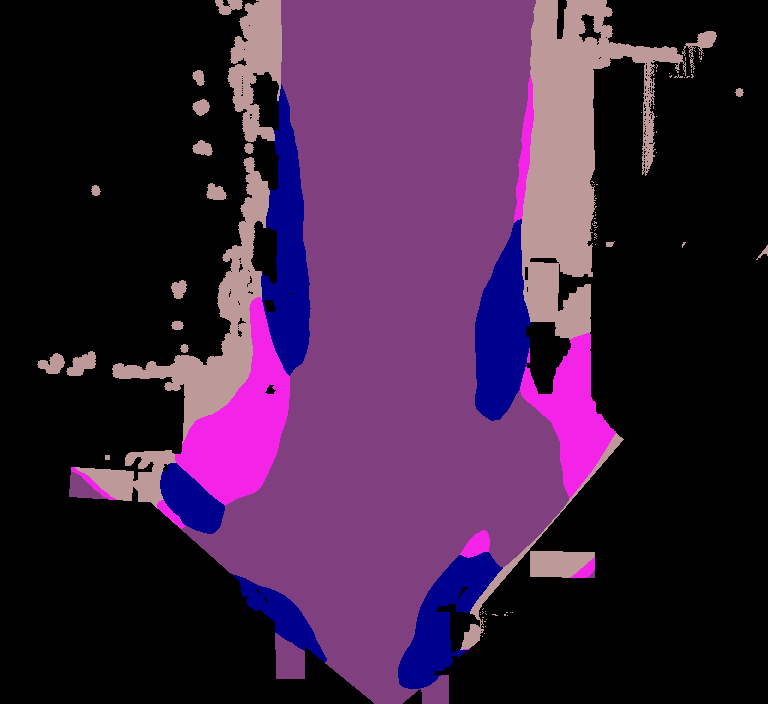} & \includegraphics[width=\linewidth, frame]{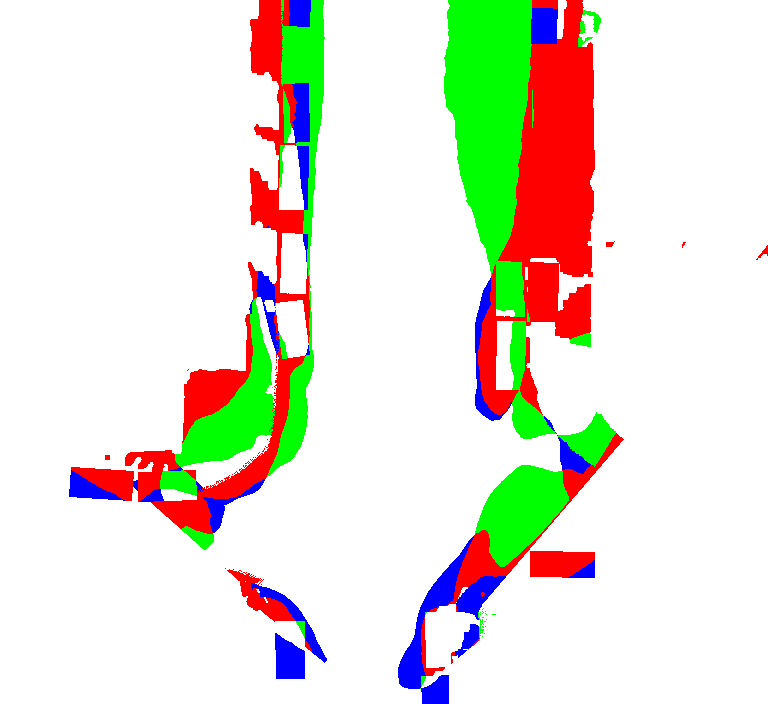} \\
\\
(f) & \includegraphics[width=\linewidth, height=0.455\linewidth, frame]{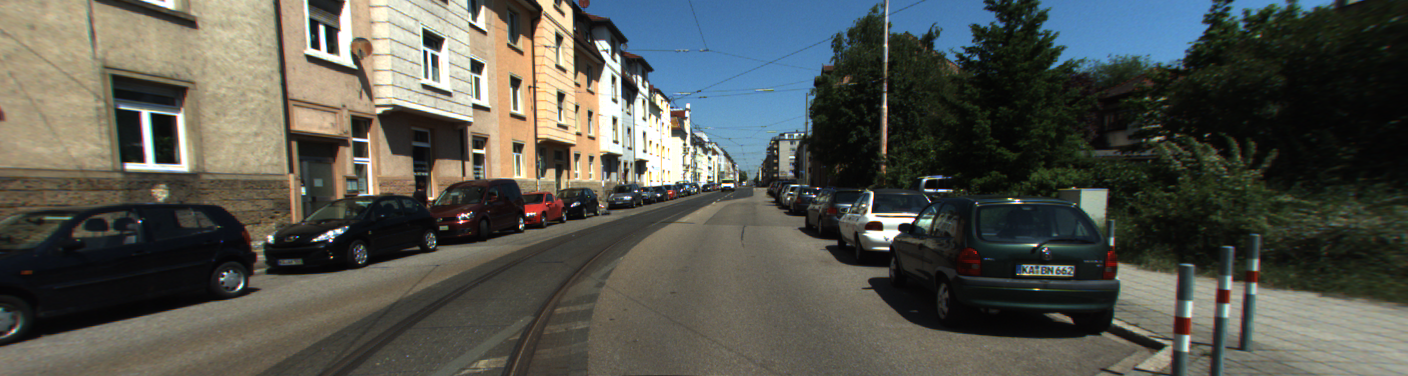} & \includegraphics[width=\linewidth, frame]{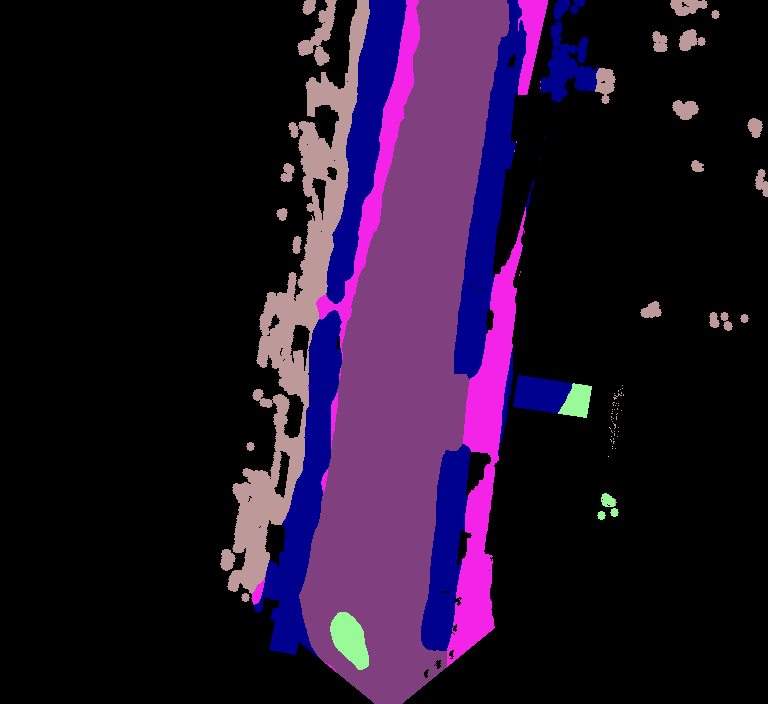} & \includegraphics[width=\linewidth, frame]{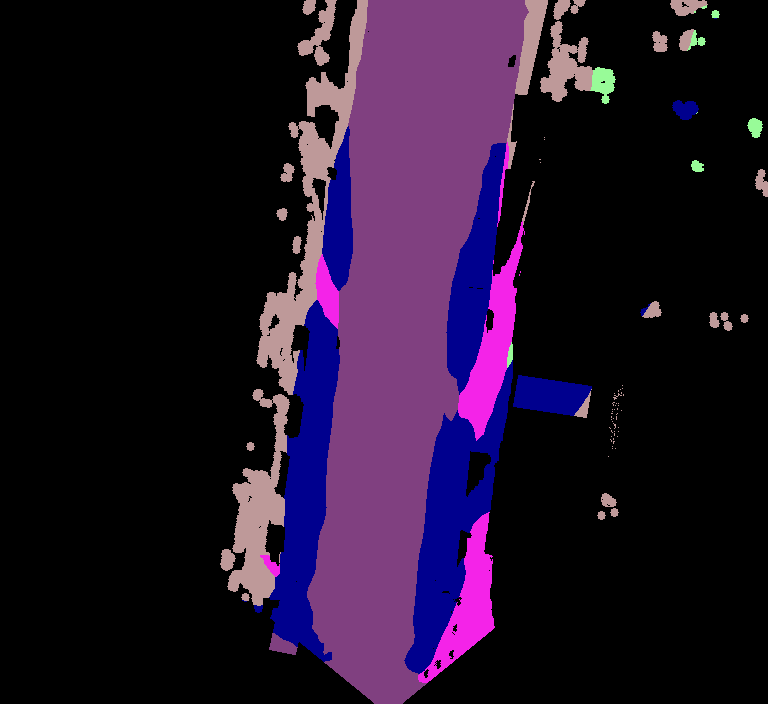} & \includegraphics[width=\linewidth, frame]{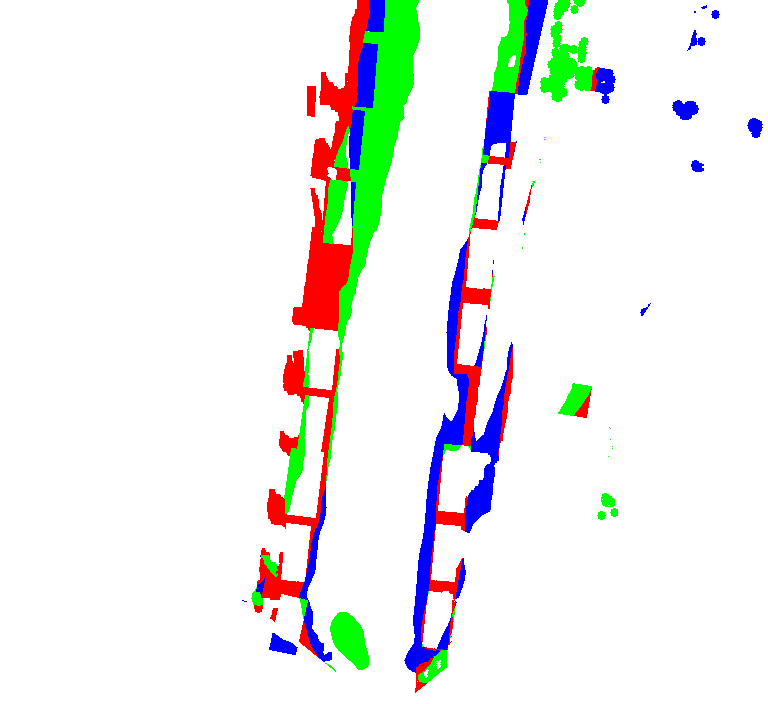} \\
\end{tabular}
}
\caption{Additional Qualitative Results on the KITTI-360 dataset. The rightmost column shows the Improvement/Error map which depicts pixels misclassified by PoBEV and correctly predicted by \net~in green, pixels misclassified by \net~and correctly by PoBEV in blue, and pixels misclassified by both models in red.}
\label{fig:supp-qual-results}
\vspace{-0.2cm}
\end{figure*}

\begin{figure*}
\centering
\footnotesize
\setlength{\tabcolsep}{0.05cm}% for the horiz padding
{
\renewcommand{\arraystretch}{0.2}% for the vertical padding
\newcolumntype{M}[1]{>{\centering\arraybackslash}m{#1}}
\begin{tabular}{cM{4.6cm}M{2.3cm}M{2.3cm}M{2.3cm}M{2.3cm}M{2.3cm}}
& Input FV Image & $0.1\%$ & $1\%$ & $10\%$ & $50\%$ & $100\%$ \\
\\
(a) & {\includegraphics[width=\linewidth, height=0.455\linewidth, frame]{images/qual_mp/rgb_kitti-1575.png}} & {\includegraphics[width=\linewidth, frame]{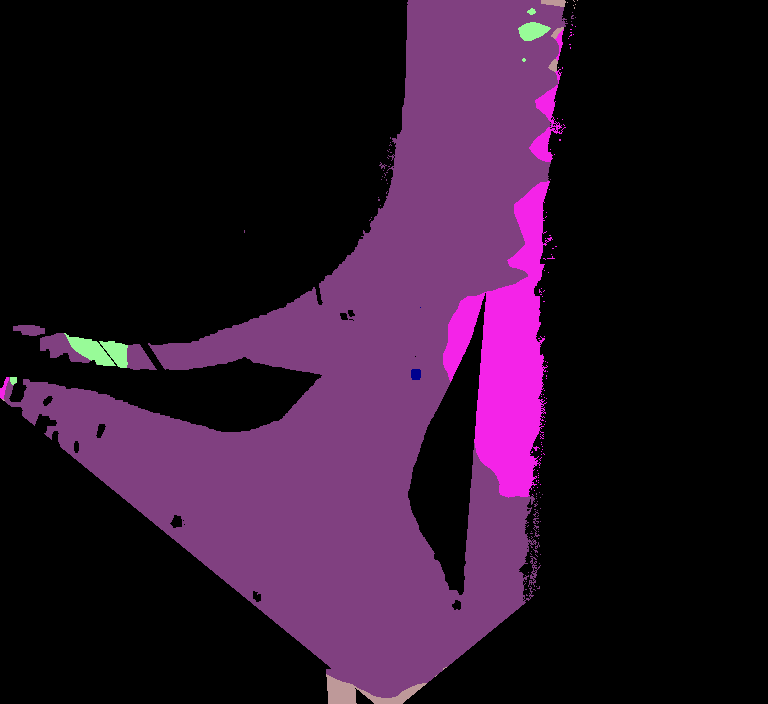}} & {\includegraphics[width=\linewidth, frame]{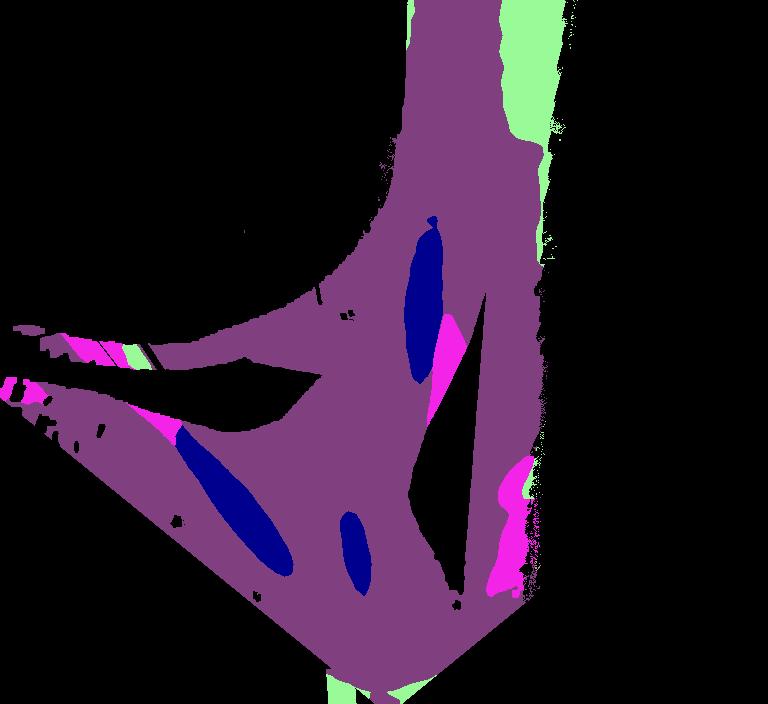}} & {\includegraphics[width=\linewidth, frame]{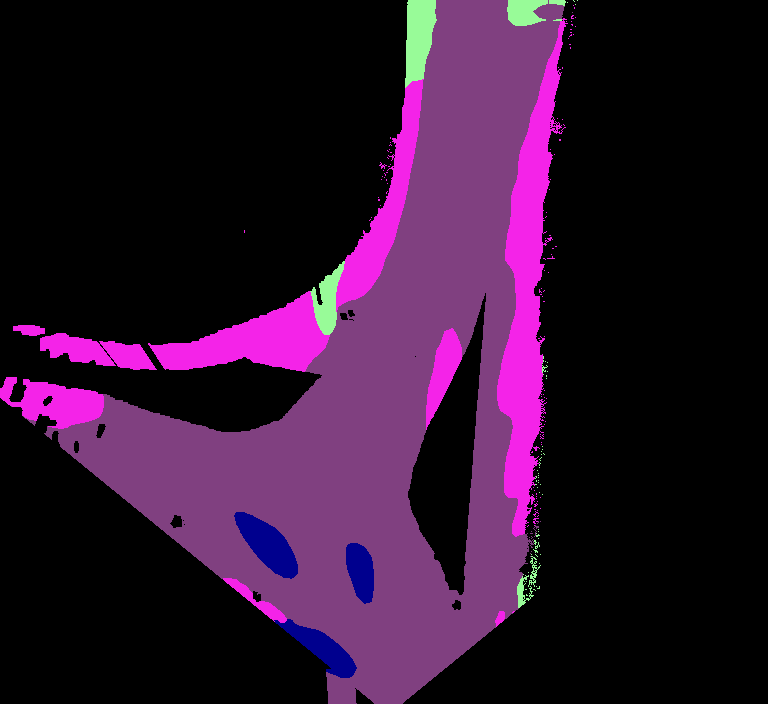}} & {\includegraphics[width=\linewidth, frame]{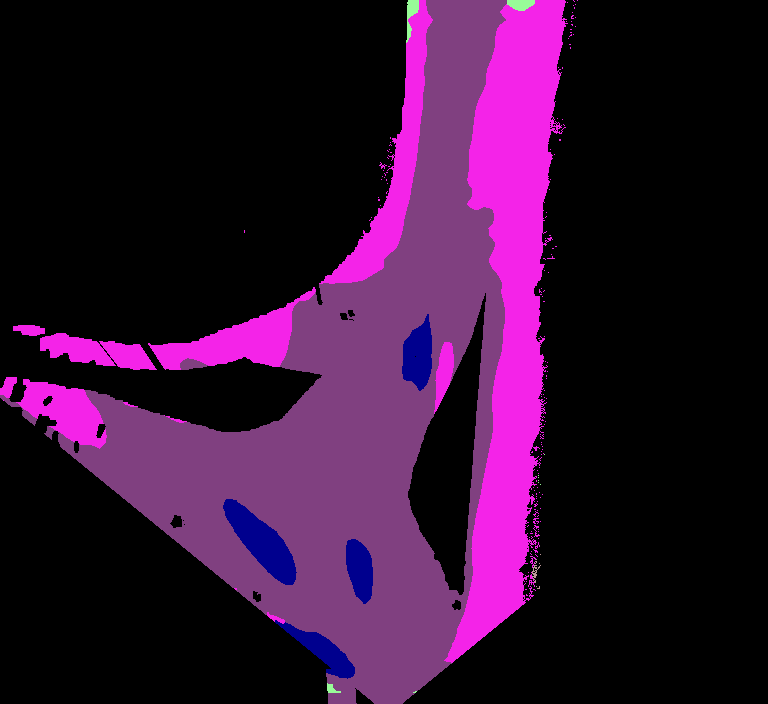}} & {\includegraphics[width=\linewidth, frame]{images/qual_mp/skyeye_kitti-1575.png}}  \\
\\
(b) & {\includegraphics[width=\linewidth, height=0.455\linewidth, frame]{images/qual_mp/rgb_kitti-0042.png}} & {\includegraphics[width=\linewidth, frame]{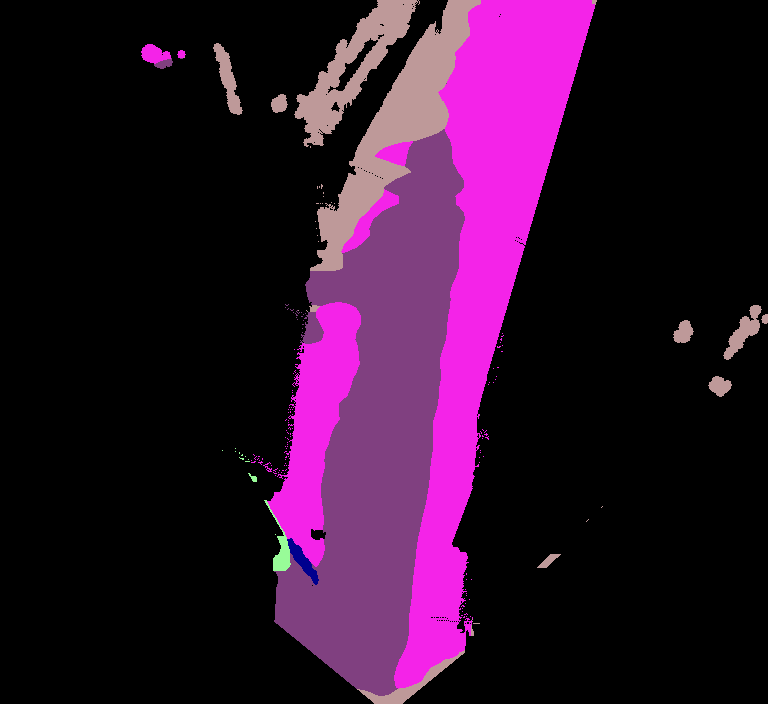}} & {\includegraphics[width=\linewidth, frame]{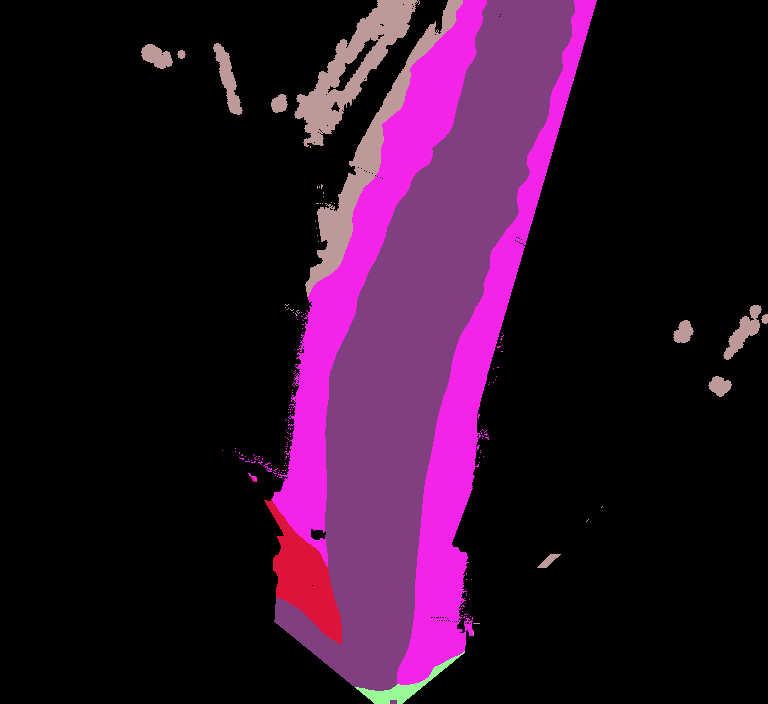}} & {\includegraphics[width=\linewidth, frame]{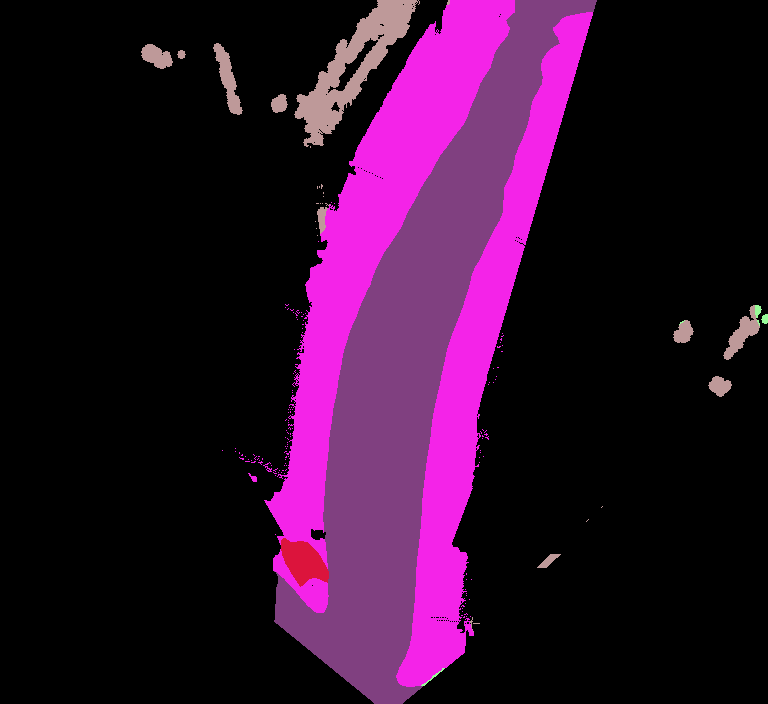}} & {\includegraphics[width=\linewidth, frame]{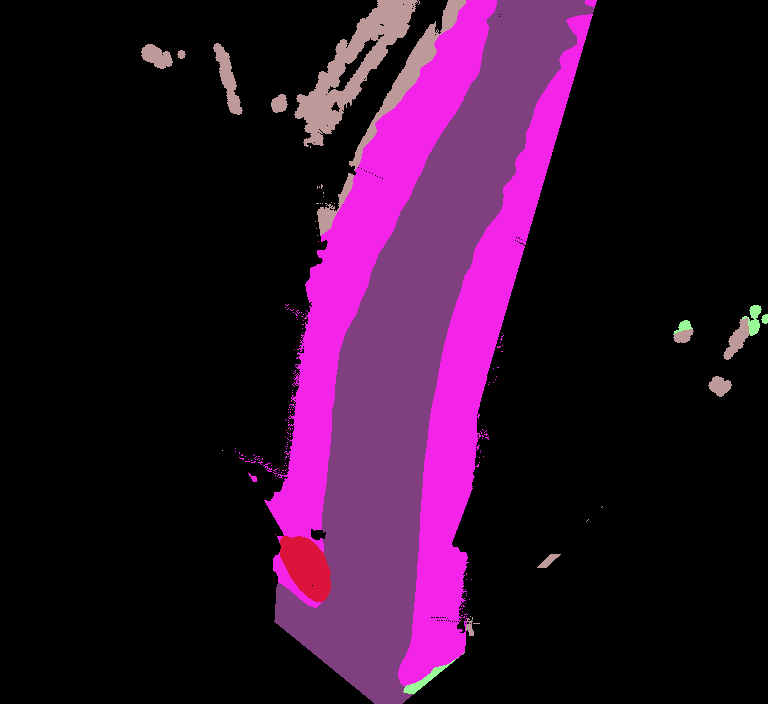}} & {\includegraphics[width=\linewidth, frame]{images/qual_mp/skyeye_kitti-0042.png}}  \\
\\
(c) & {\includegraphics[width=\linewidth, height=0.455\linewidth, frame]{images/qual_mp/rgb_kitti-0179.png}} & {\includegraphics[width=\linewidth, frame]{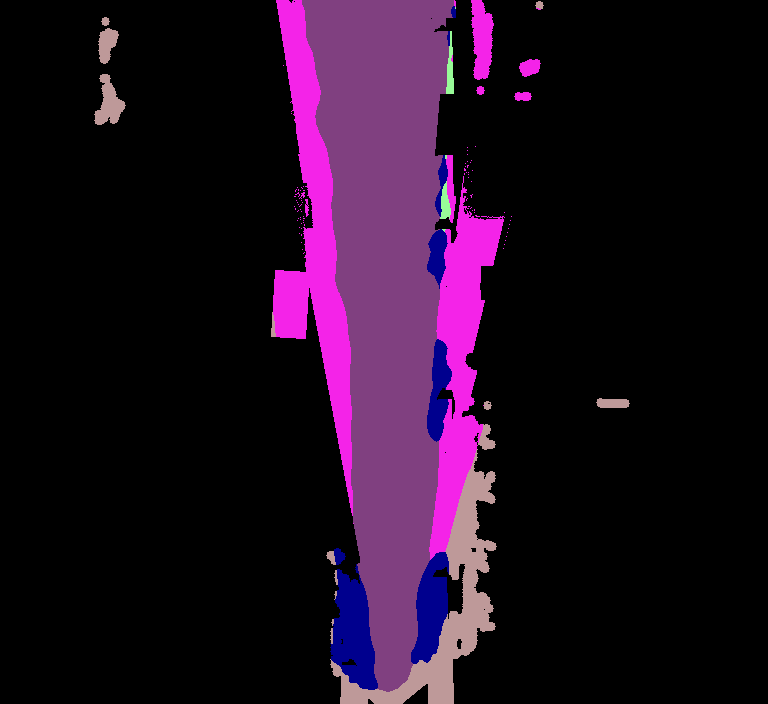}} & {\includegraphics[width=\linewidth, frame]{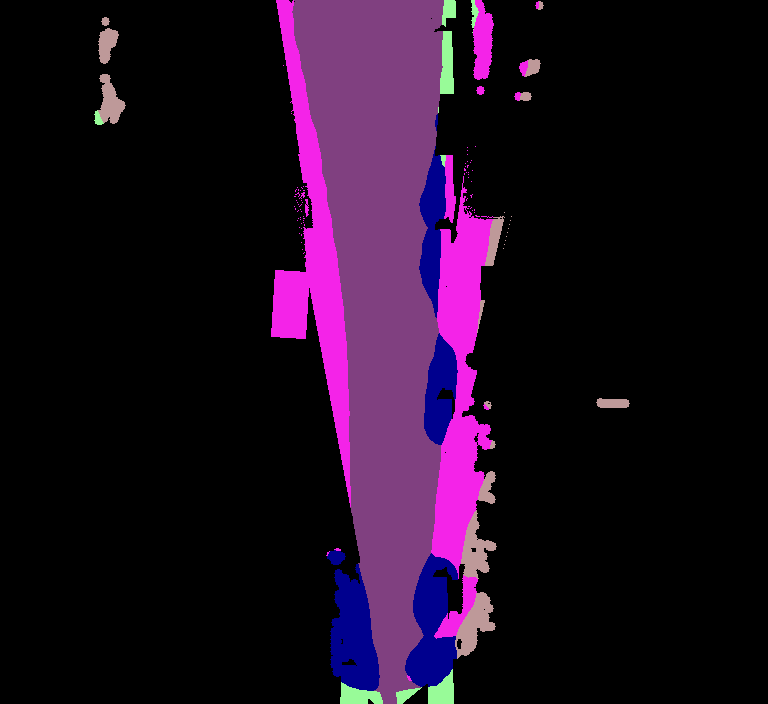}} & {\includegraphics[width=\linewidth, frame]{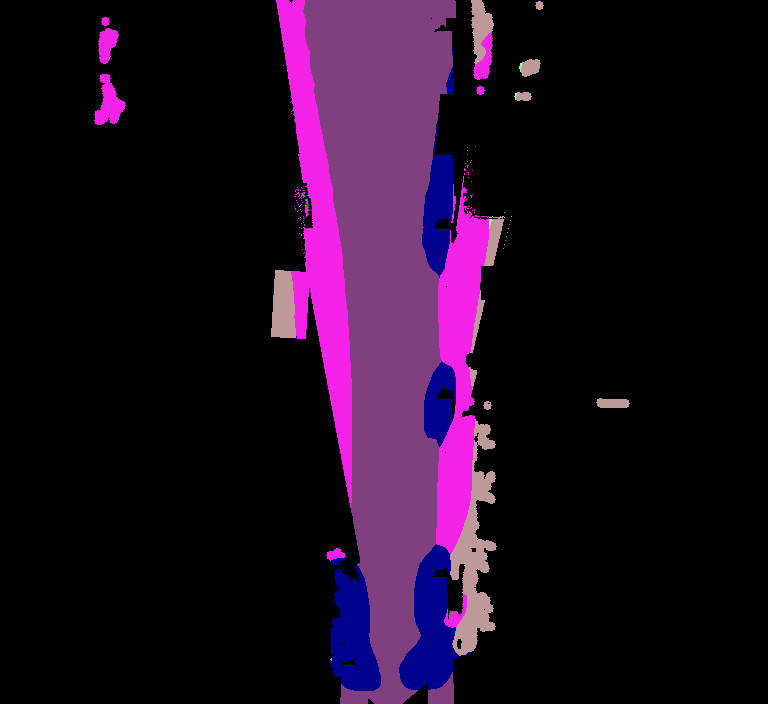}} & {\includegraphics[width=\linewidth, frame]{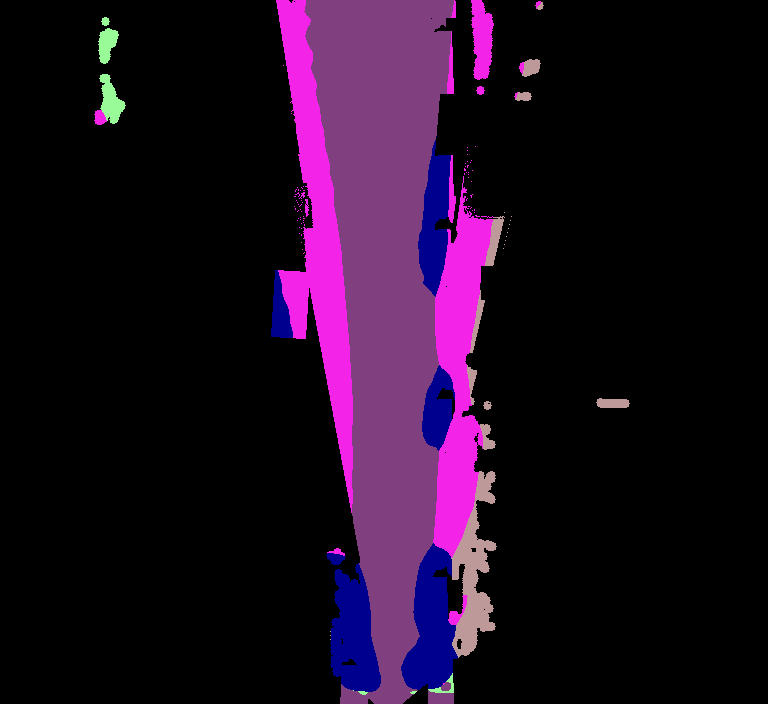}} & {\includegraphics[width=\linewidth, frame]{images/qual_mp/skyeye_kitti-0179.png}} \\
\\
(d) & {\includegraphics[width=\linewidth, height=0.455\linewidth, frame]{images/qual_mp/rgb_kitti-0212.png}} & {\includegraphics[width=\linewidth, frame]{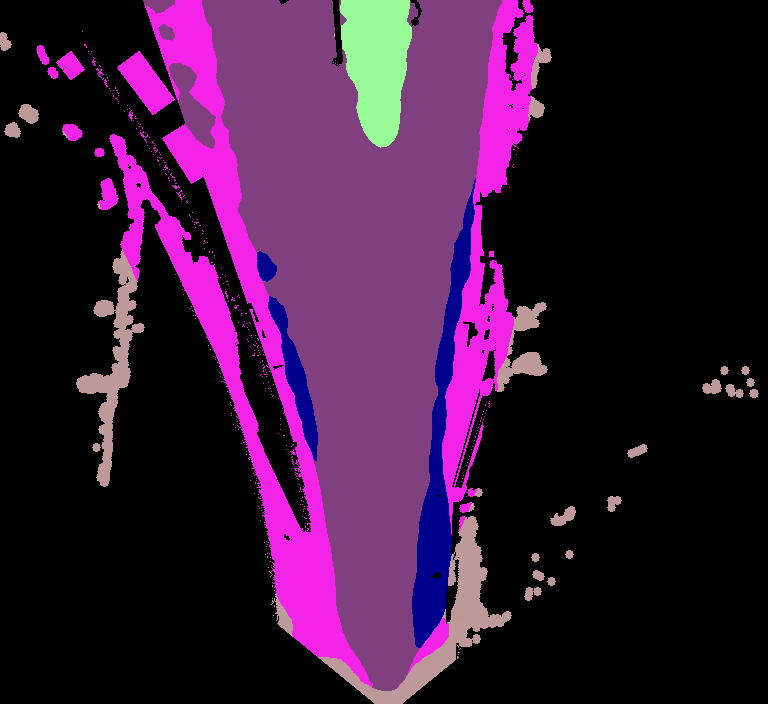}} & {\includegraphics[width=\linewidth, frame]{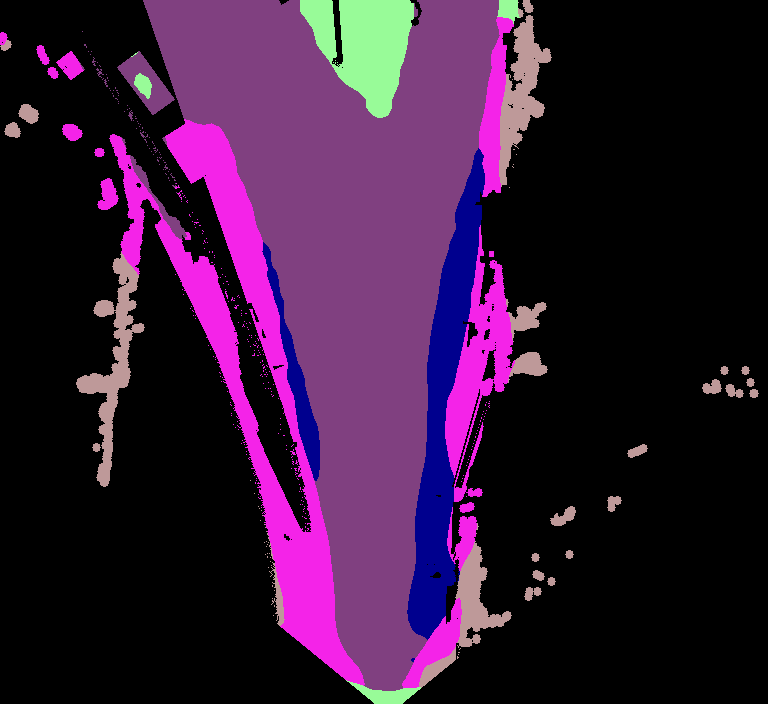}} & {\includegraphics[width=\linewidth, frame]{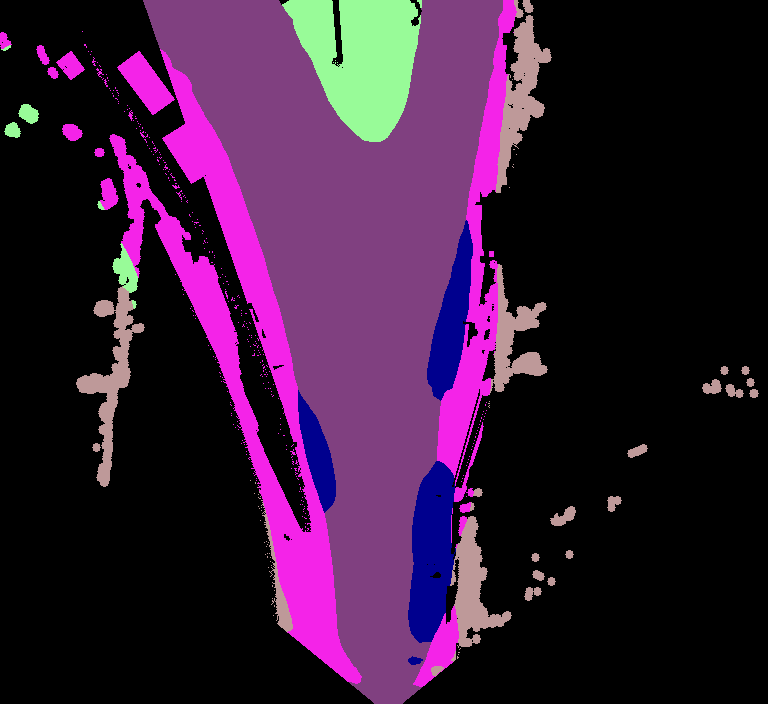}} & {\includegraphics[width=\linewidth, frame]{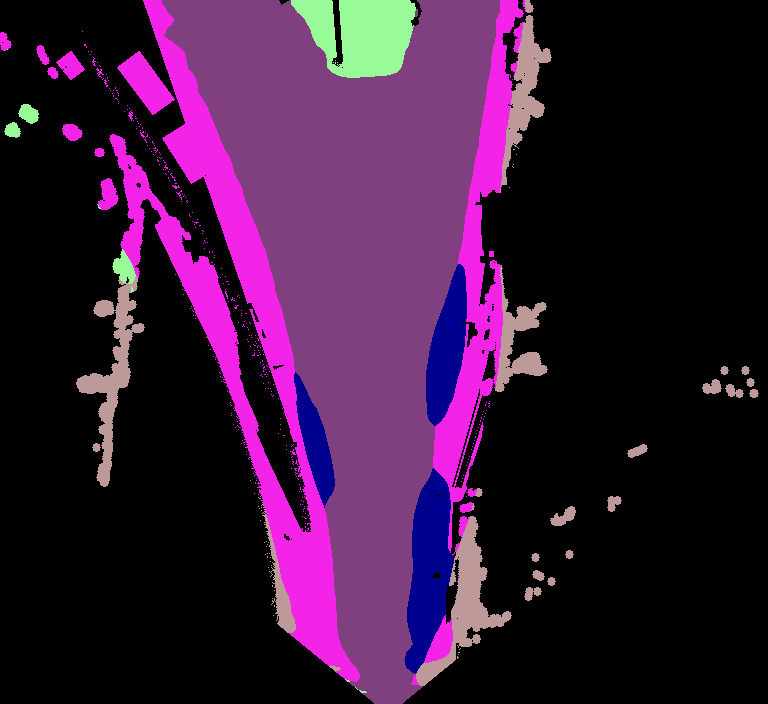}} & {\includegraphics[width=\linewidth, frame]{images/qual_mp/skyeye_kitti-0212.png}}  \\
\\
(e) & {\includegraphics[width=\linewidth, height=0.455\linewidth, frame]{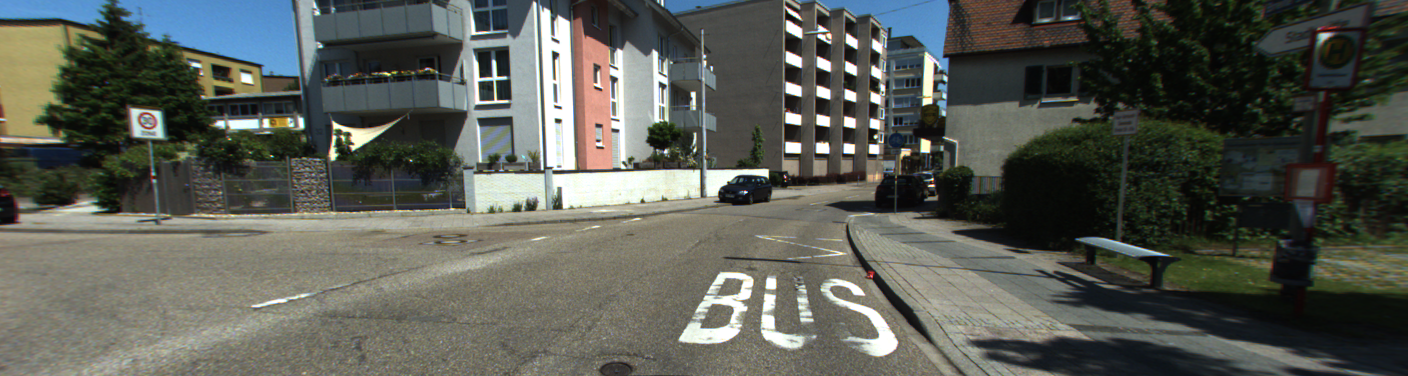}} & {\includegraphics[width=\linewidth, frame]{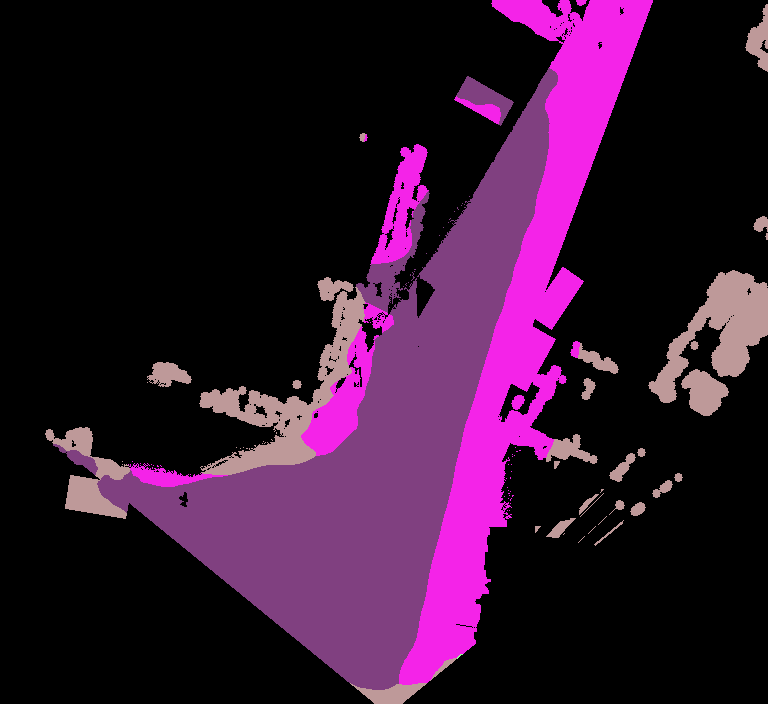}} & {\includegraphics[width=\linewidth, frame]{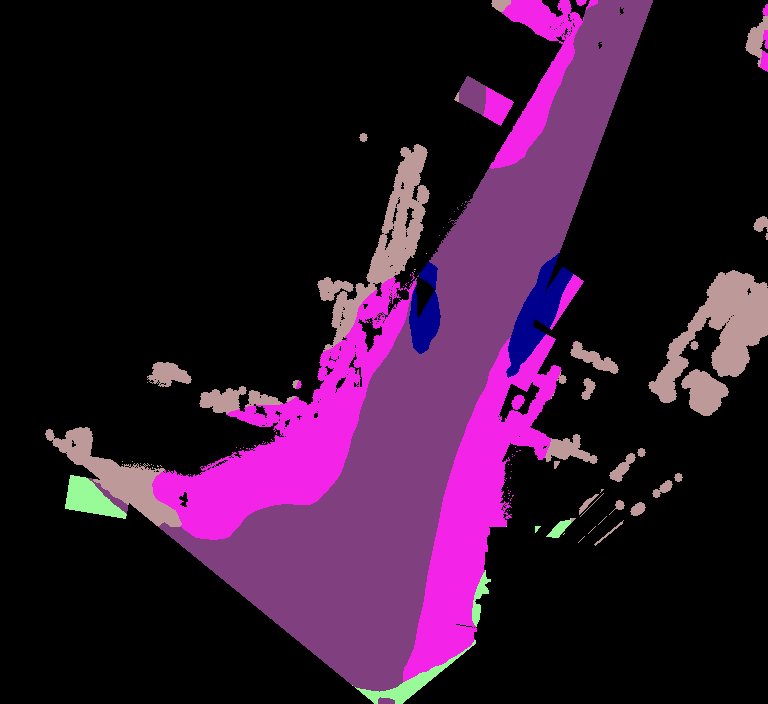}} & {\includegraphics[width=\linewidth, frame]{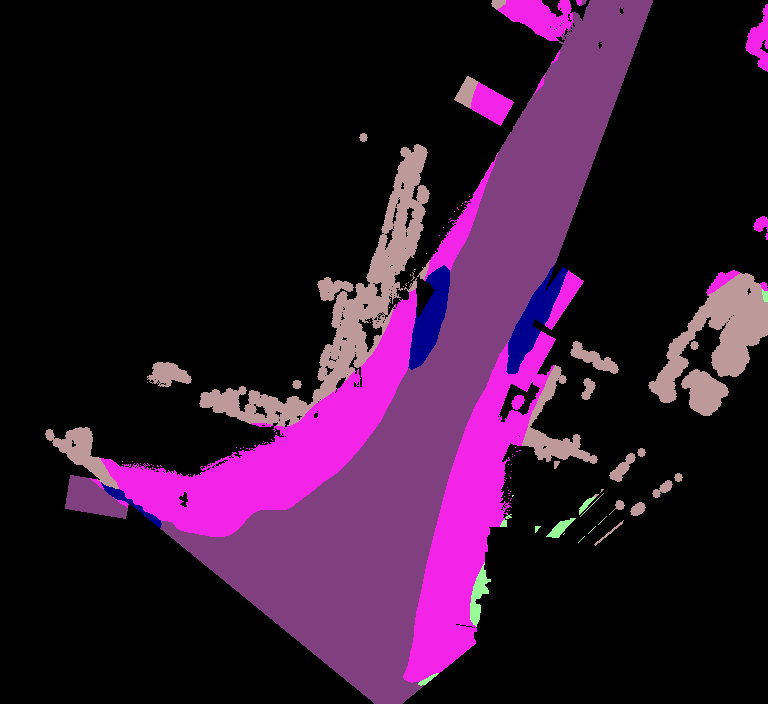}} & {\includegraphics[width=\linewidth, frame]{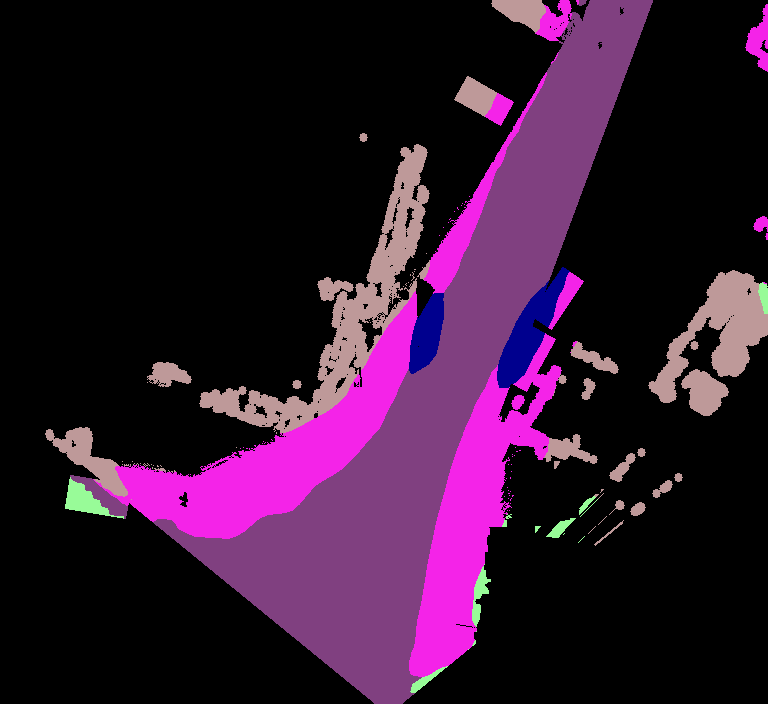}} & {\includegraphics[width=\linewidth, frame]{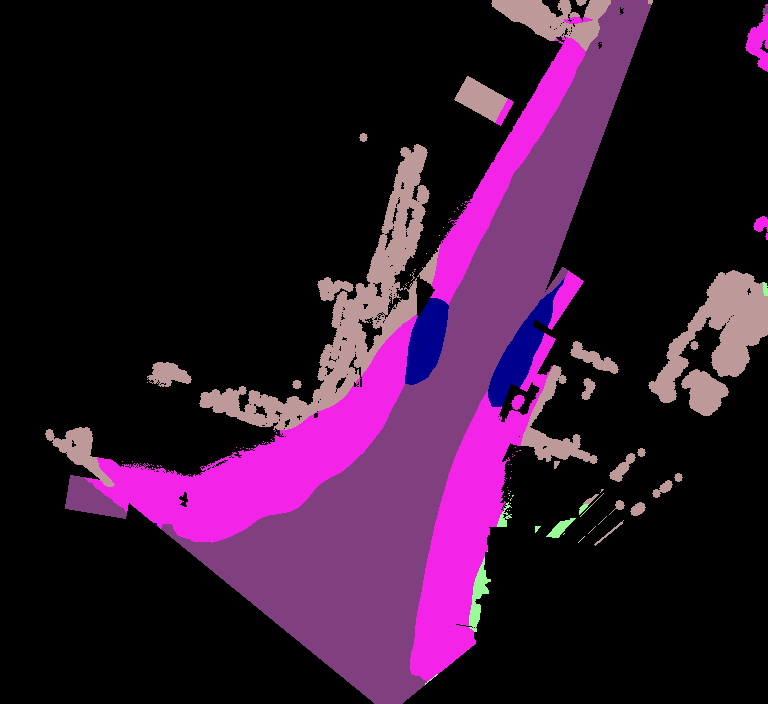}}  \\
\\
(f) & {\includegraphics[width=\linewidth, height=0.455\linewidth, frame]{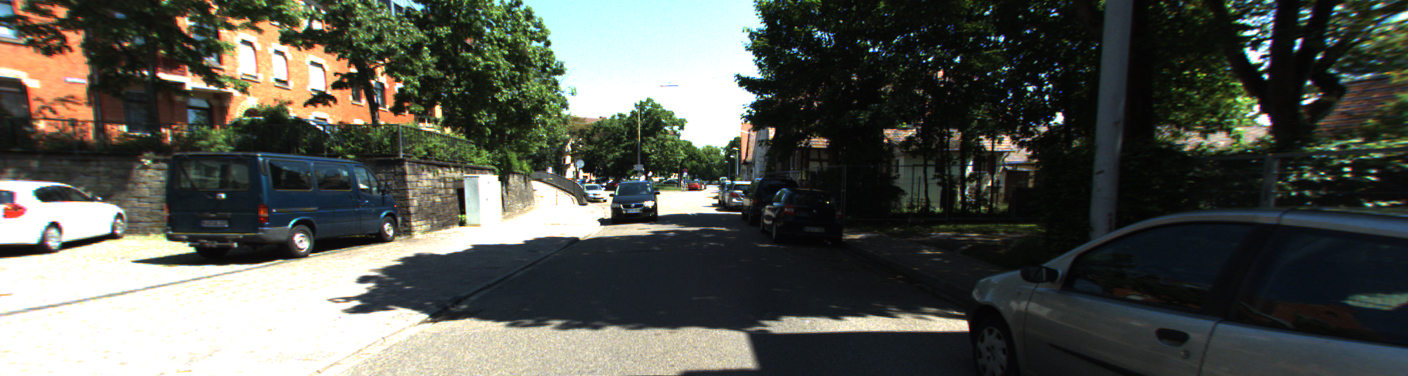}} & {\includegraphics[width=\linewidth, frame]{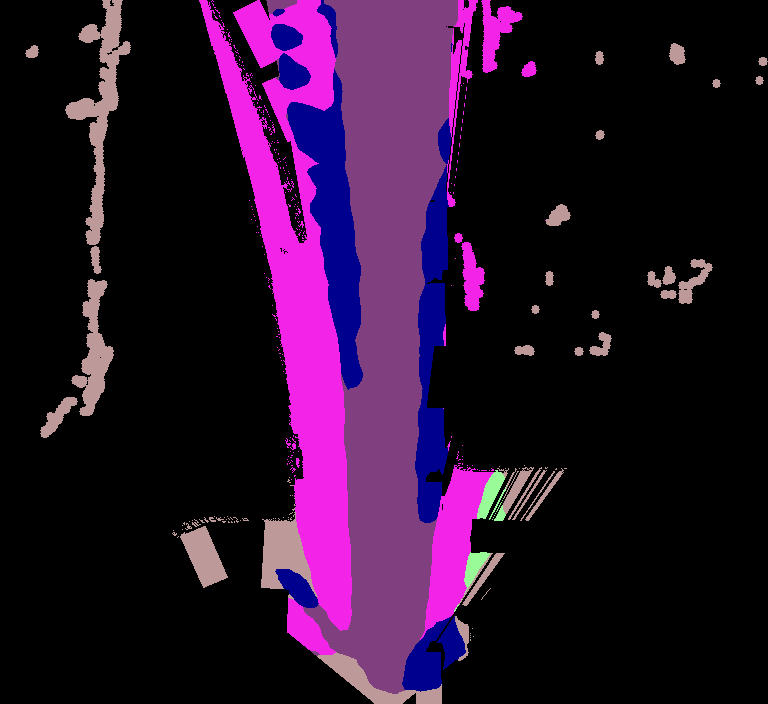}} & {\includegraphics[width=\linewidth, frame]{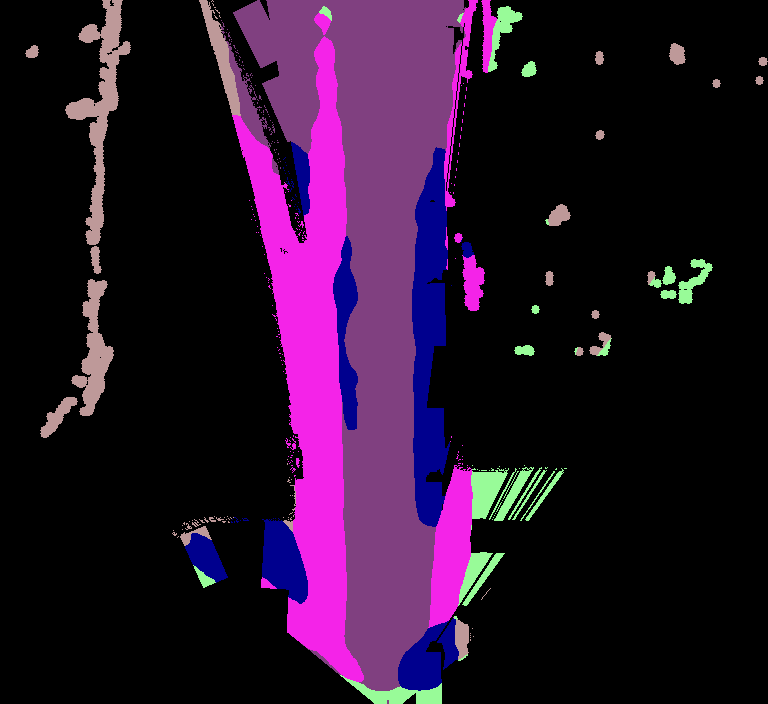}} & {\includegraphics[width=\linewidth, frame]{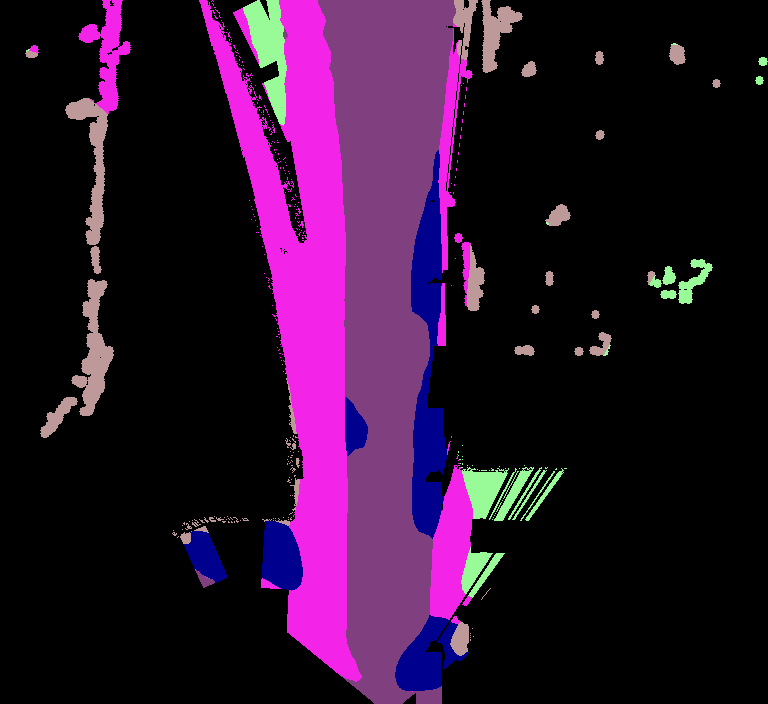}} & {\includegraphics[width=\linewidth, frame]{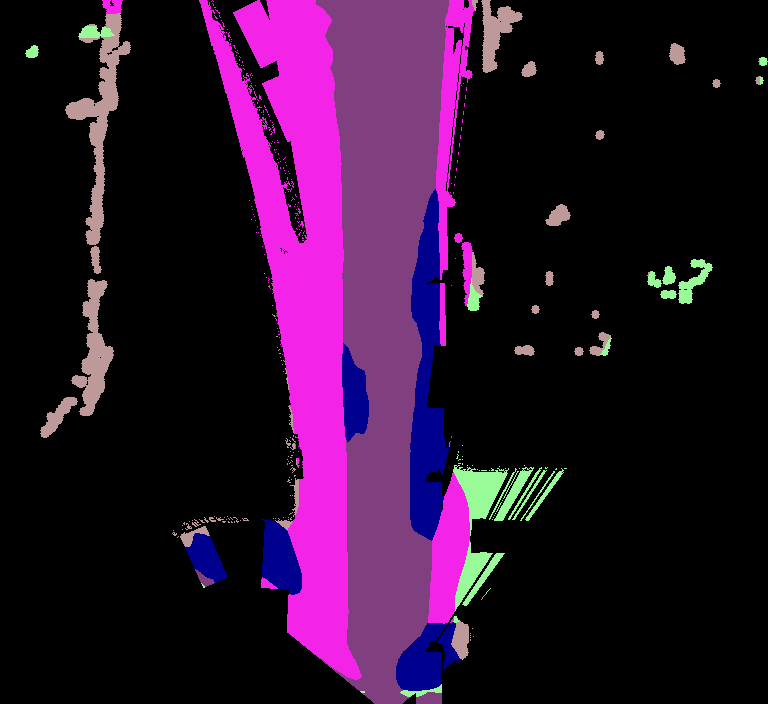}} & {\includegraphics[width=\linewidth, frame]{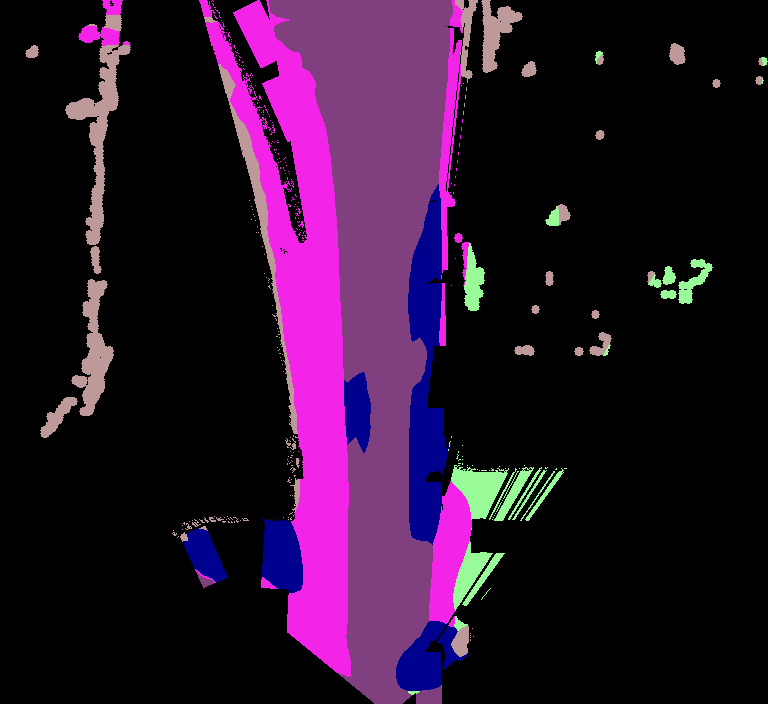}}  \\
\\
(g) & {\includegraphics[width=\linewidth, height=0.455\linewidth, frame]{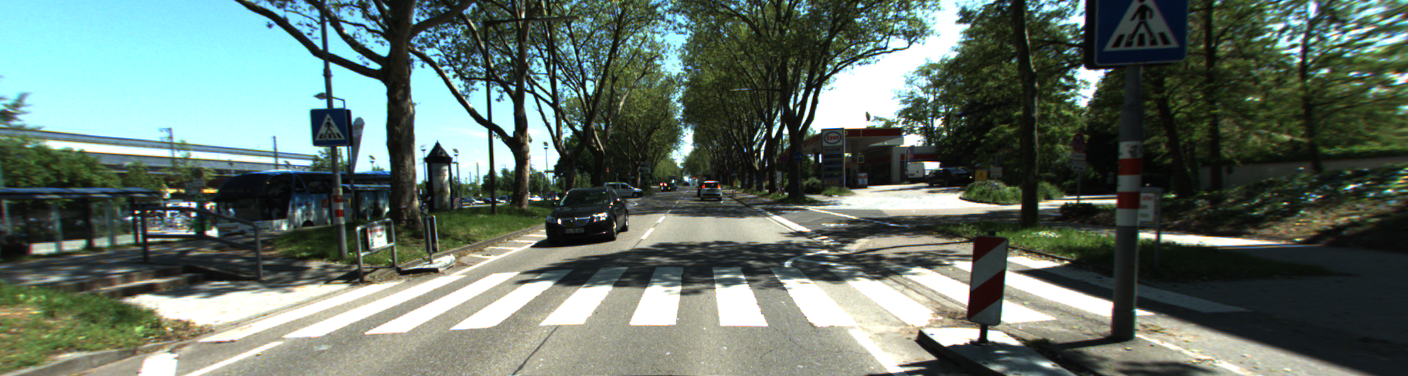}} & {\includegraphics[width=\linewidth, frame]{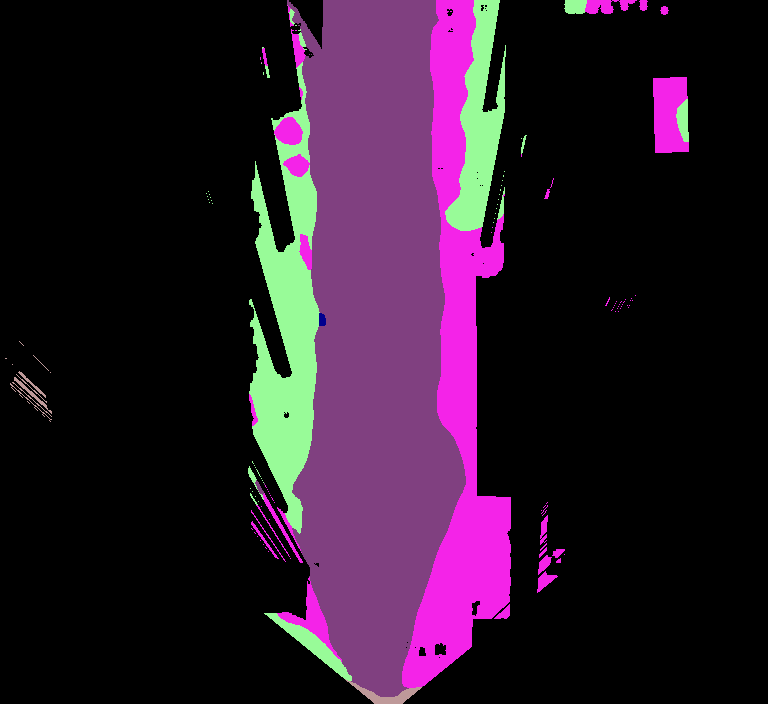}} & {\includegraphics[width=\linewidth, frame]{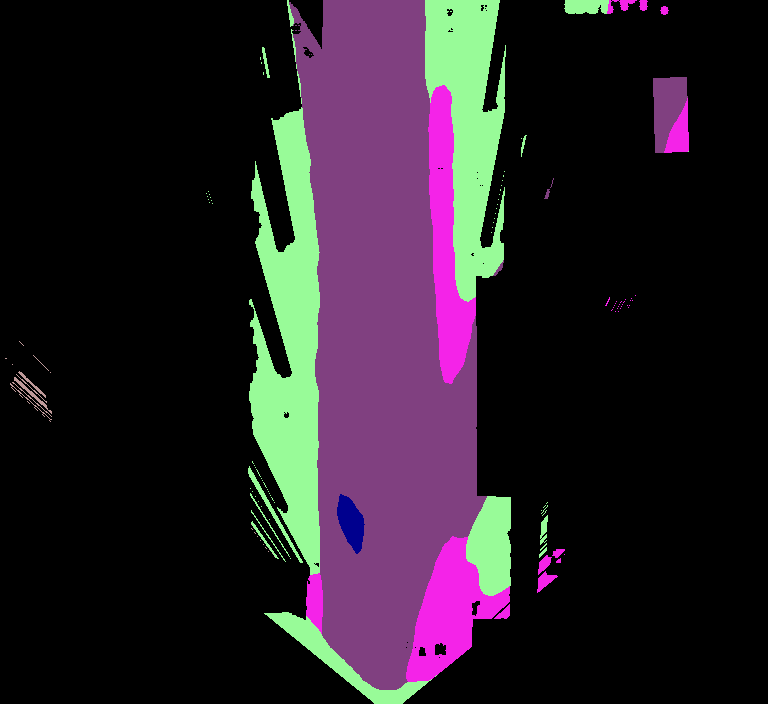}} & {\includegraphics[width=\linewidth, frame]{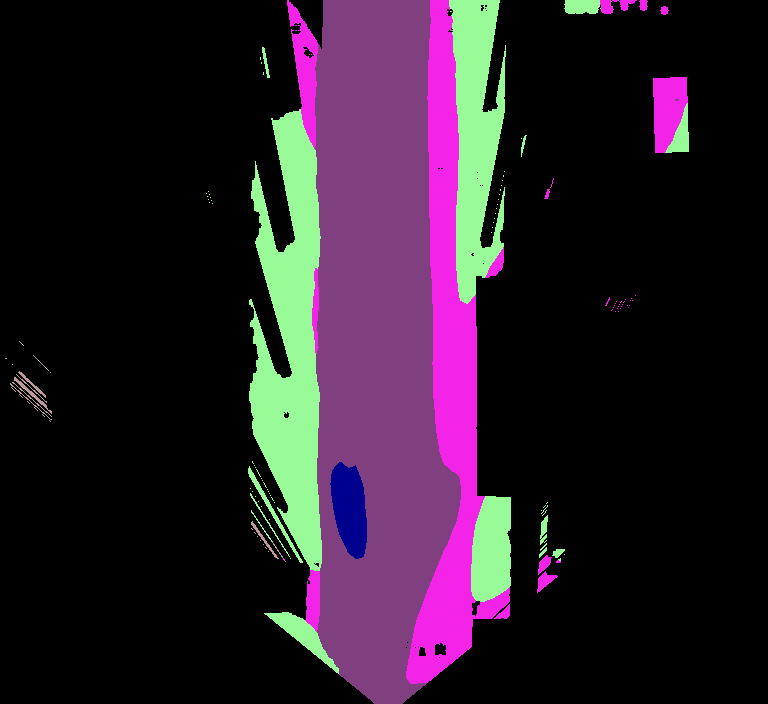}} & {\includegraphics[width=\linewidth, frame]{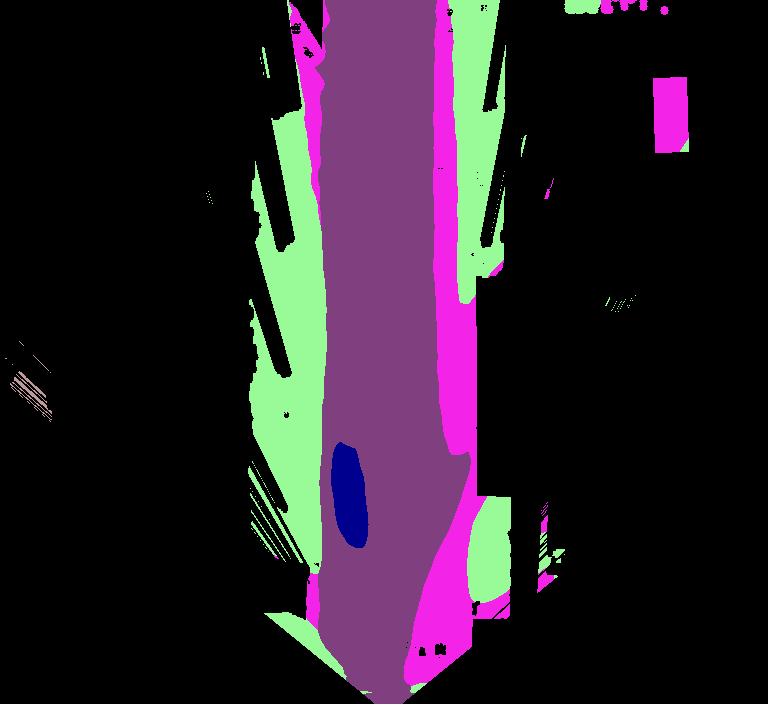}} & {\includegraphics[width=\linewidth, frame]{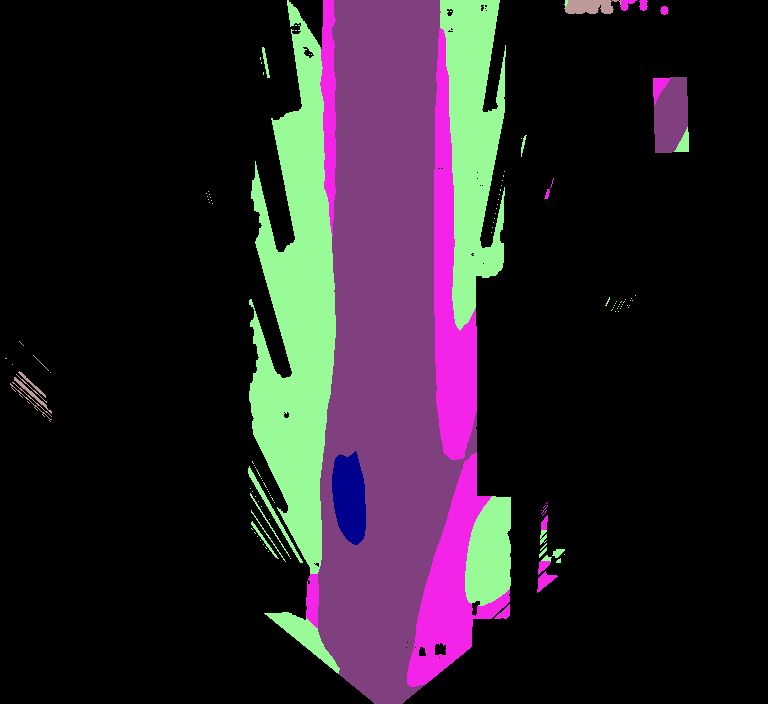}}  \\
\\
(h) & {\includegraphics[width=\linewidth, height=0.455\linewidth, frame]{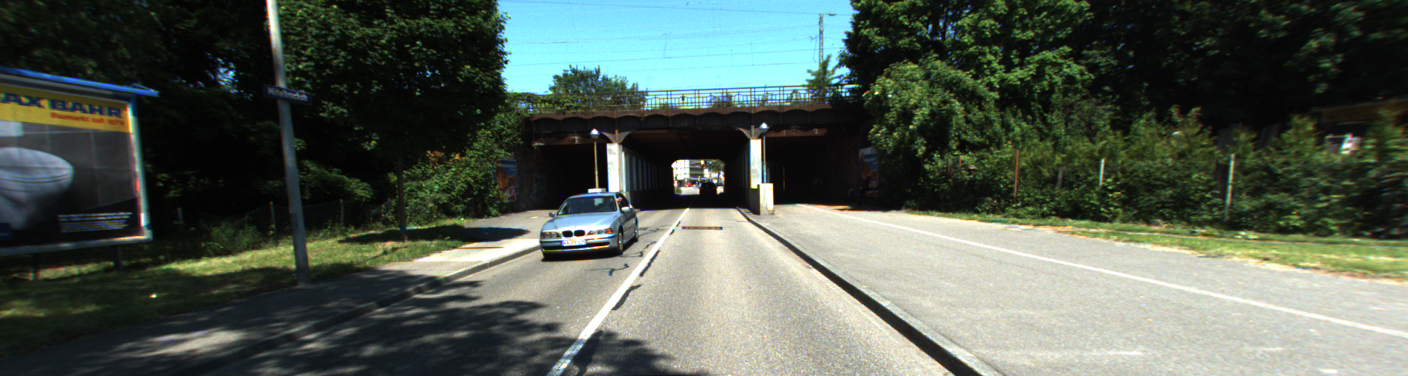}} & {\includegraphics[width=\linewidth, frame]{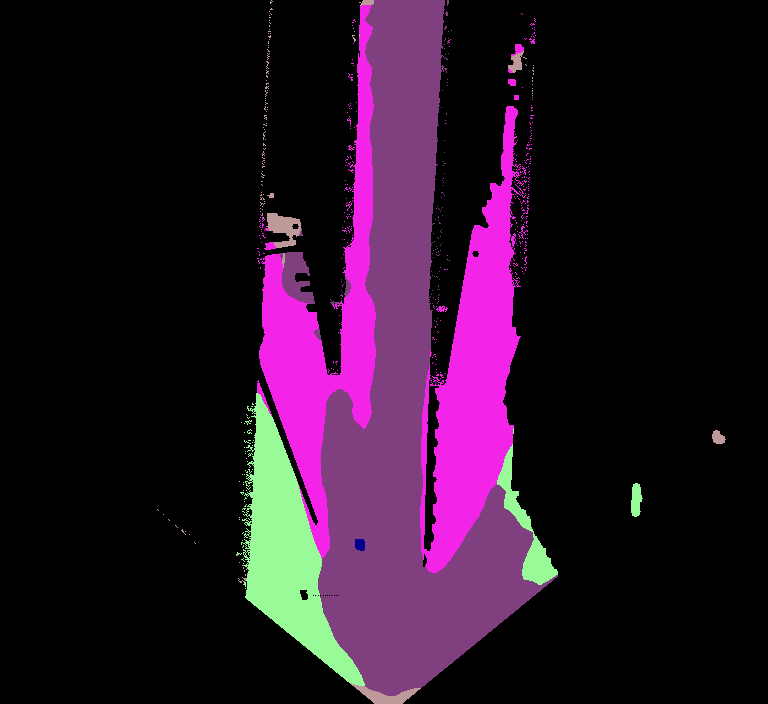}} & {\includegraphics[width=\linewidth, frame]{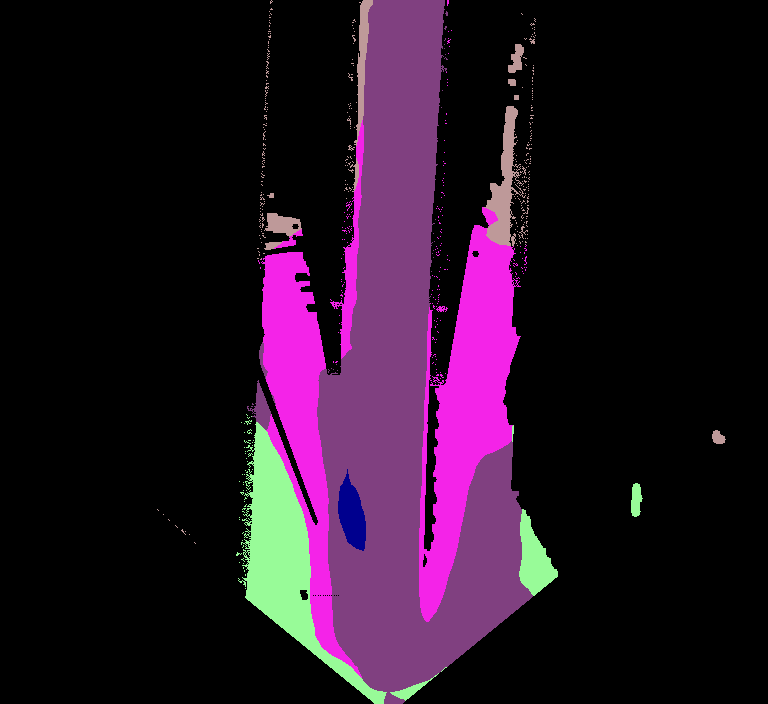}} & {\includegraphics[width=\linewidth, frame]{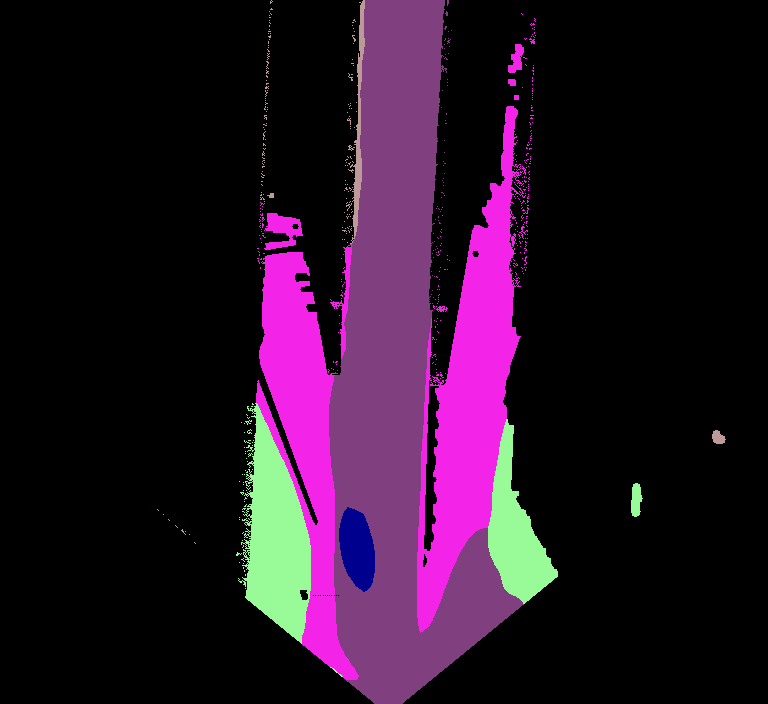}} & {\includegraphics[width=\linewidth, frame]{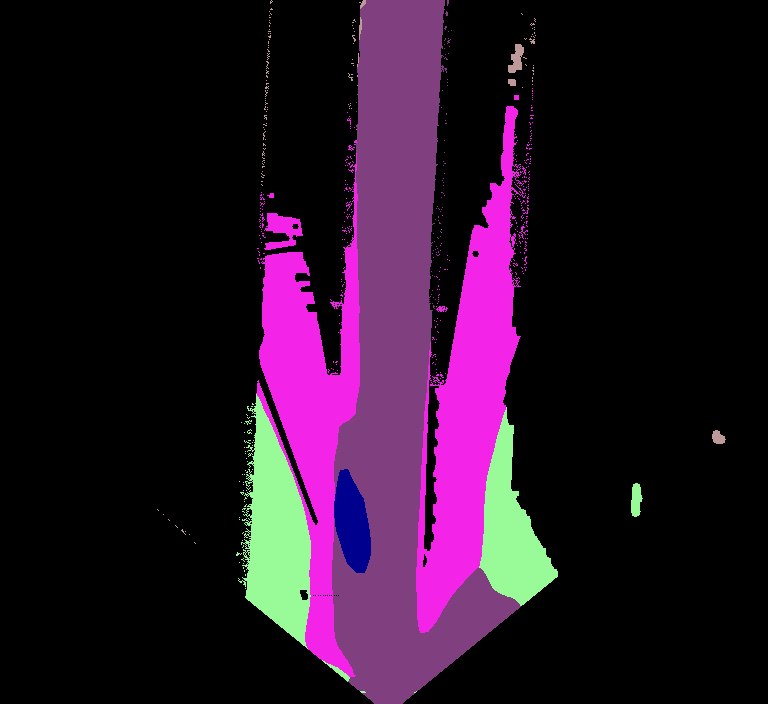}} & {\includegraphics[width=\linewidth, frame]{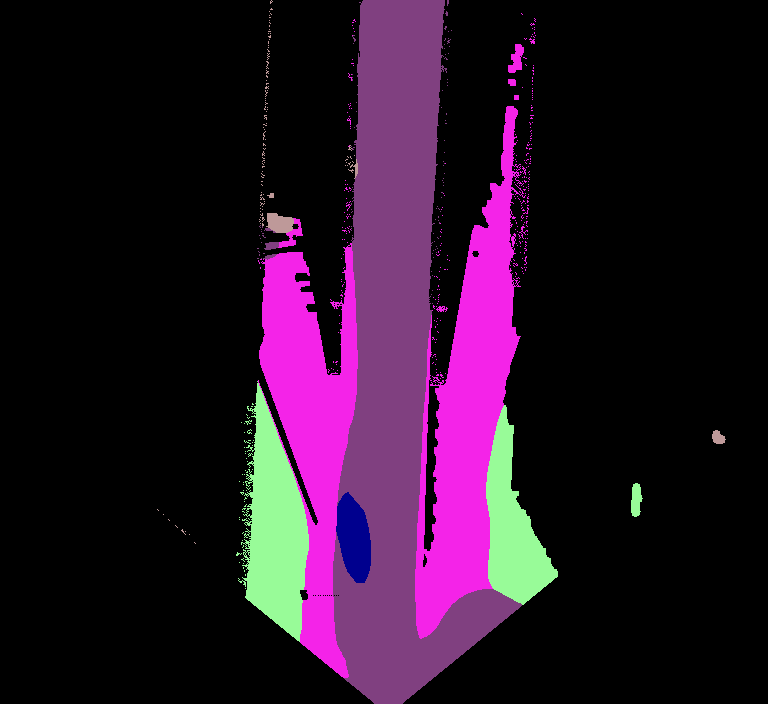}}  \\
\end{tabular}
}
\caption{Qualitative results obtained when \net~is trained using $0.1\%$, $1\%$, $10\%$, $50\%$ and $100\%$ of BEV pseudolabels.}
\label{fig:supp-percentage-results}
\vspace{-0.2cm}
\end{figure*}

We qualitatively demonstrate the performance of our model by comparing the output from our network to those obtained from the state-of-the-art fully supervised approach, PoBEV~\cite{cit:bev-seg-panopticbev}. We also show an Error/Improvement map in the rightmost column to highlight the difference in predictions between both approaches. \cref{fig:supp-qual-results} presents the semantic BEV map predictions obtained from both networks. We observe from the figures that our approach is mostly on par with PoBEV across a wide range of scenarios from straight roads with multiple parked cars, to curved roads, and to complex intersections. As already noted in the main paper, we observe from \cref{fig:supp-qual-results}(d, e, f) that \net~performs significantly better on static regions as compared to PoBEV which can be confirmed by looking at the large swathes of green in the last column. Our model is able to estimate the extent of roads and sidewalks better than PoBEV, and this improvement in performance can be attributed to training using implicit supervision which encourages the model to learn spatially coherent representations. We also observe from \cref{fig:supp-qual-results}(a, b, c) that our model is able to accurately estimate the locations of multiple cars in the scene and these images qualitatively look very similar to that of PoBEV.
\clearpage
\noindent However, we note that when cars are extremely close to each other, our model sometimes stretches the extent of cars and merges multiple cars into one big blob. This is a limitation of our model and is largely a consequence of using only a forward-facing camera during implicit supervision which inhibits the model from extracting the extents of all four sides of vehicles. Furthermore, our method does not have access to the ground truth BEV map obtained using LiDAR data that is able to perceive both close and distant vehicles with higher precision. Nevertheless, our model performs on par with PoBEV without using any ground truth supervision in BEV, thus highlighting the impact of our self-supervised BEV semantic mapping framework.

\subsection{Different Percentages of BEV Pseudolabels}
In this section, we qualitatively evaluate the impact of using different percentages of BEV pseudo labels on the overall performance of our model. \cref{fig:supp-percentage-results} presents the results of this evaluation. We observe a very interesting trend across samples when training our model with different percentages of BEV pseudo labels. Across all images in \cref{fig:supp-percentage-results}, we observe that our model gradually improves its reasoning about dynamic cars in the scene with an increase in the percentage of BEV pseudo labels. When using only $0.1\%$ of pseudo labels, our model fails to identify any dynamic cars in the scene. Upon increasing the number of pseudo labels to $1\%$, we see that our model starts reasoning about the locations of dynamic objects but the cars look stretched and artificial. However, upon further increasing the percentage of BEV pseudo labels to $10\%$, our model predicts the locations as well as the extent of cars accurately. Further increase in the percentage of pseudo labels refines the predictions and better constrains the extent of vehicles, but no significant improvement can be observed between the BEV maps obtained using $10\%$, $50\%$ and $100\%$ of pseudo labels. This supports our findings in Tab. 2 of the main paper where we observe no large changes in mIoU scores when using $10\%$, $50\%$, and $100\%$ of BEV pseudo labels to train the model. 

Further, a special observation can be made when the scene is fully static. \cref{fig:supp-percentage-results}(c, d, f) depict static scenes with multiple parked cars wherein we observe that our model with $0.1\%$ of pseudo labels is already able to predict the locations of cars with high accuracy. This accurate estimation of the location of static cars with very few pseudo labels can be attributed to the structure infused into the 3D voxel grid representation by implicit supervision. This strong structural signal thus enables the model to reason about the world in BEV even when the model is exposed to extremely sparse samples in BEV.

\end{document}